\pdfoutput=1

\documentclass[11pt]{article}

\usepackage[final]{acl}

\usepackage{times}
\usepackage{latexsym}
\usepackage{soul}
\usepackage{bigstrut}
\usepackage{url}
\usepackage{algorithm}
\usepackage{algorithmic}
\usepackage{amsmath,amssymb,amsfonts}
\usepackage{amsthm}
\usepackage{float}
\usepackage{enumitem}
\usepackage{booktabs}
\usepackage{amssymb}
\usepackage{colortbl}
\usepackage{graphicx}
\usepackage{multirow}
\usepackage[switch]{lineno}
\usepackage{bbding}
\usepackage{booktabs}
\usepackage{soul}
\usepackage{subfig}
\usepackage{newfloat}
\usepackage{listings}
\newcommand{\blue}[1]{\textcolor{blue}{#1}}

\usepackage[T1]{fontenc}

\usepackage[utf8]{inputenc}

\usepackage{microtype}

\usepackage{inconsolata}

\usepackage{graphicx}

\title{Efficient and Effective Prompt Tuning via Prompt Decomposition and Compressed Outer Product}

\author{
 \textbf{Pengxiang Lan\textsuperscript{1}}\thanks{The first two authors contributed equally.},
 \textbf{Haoyu Xu\textsuperscript{1}}\footnotemark[1],
 \textbf{Enneng Yang\textsuperscript{1}},
 \textbf{Yuliang Liang\textsuperscript{1}},
\\
 \textbf{Guibing Guo\textsuperscript{1}}\thanks{Corresponding author.},
 \textbf{Jianzhe Zhao\textsuperscript{1}},
 \textbf{Xingwei Wang\textsuperscript{2}}
\\
\textsuperscript{1}Software College, Northeastern University, China,
\\
\textsuperscript{2}School of Computer Science and Engineering, Northeastern University, China
\\
\{pengxianglan, haoyuxu, ennengyang, liangyuliang\}@stumail.neu.edu.cn, 
\\
\{guogb, zhaojz\}@swc.neu.edu.cn, wangxw@mail.neu.edu.cn
}

\begin{document}
\maketitle
\begin{abstract}
Prompt tuning (PT) offers a cost-effective alternative to fine-tuning large-scale pre-trained language models (PLMs), requiring only a few parameters in soft prompt tokens added before the input text. However, existing PT approaches face two significant issues: (\textbf{i}) They overlook intrinsic semantic associations between soft prompt tokens, leading to high discreteness and limited interactions, thus reducing the model's comprehension and effectiveness in complex tasks.
(\textbf{ii}) Due to the complexity of downstream tasks, long soft prompt is necessitated to improve performance, but prompt length correlates positively with memory usage and computational costs. Achieving high efficiency and performance remains an ongoing challenge. To address these issues,  we propose a novel \textbf{L}ow-p\textbf{A}ra\textbf{M}eters \textbf{P}rompt Tuning (\textbf{LAMP}) method, which leverages prompt decomposition and compressed outer product.
Specifically, the prompt decomposition module employs Truncated SVD to reduce training parameters and significantly lower the dimensionality of the soft prompt parameter space. It then utilizes a compressed outer product module to facilitate multiple interactions among prompt tokens, exploring their intrinsic associations to enhance knowledge representation. Finally, LAMP uses average pooling to reduce memory usage and training/inference time. Extensive experiments across six architectures and eight datasets demonstrate that LAMP outperforms state-of-the-art PT-based and LoRA-based methods in performance and efficiency. 
\end{abstract}

\begin{figure}[t]
\centering
  \subfloat{
  \includegraphics[scale=0.188, trim={0mm 0mm 0mm 0mm}]{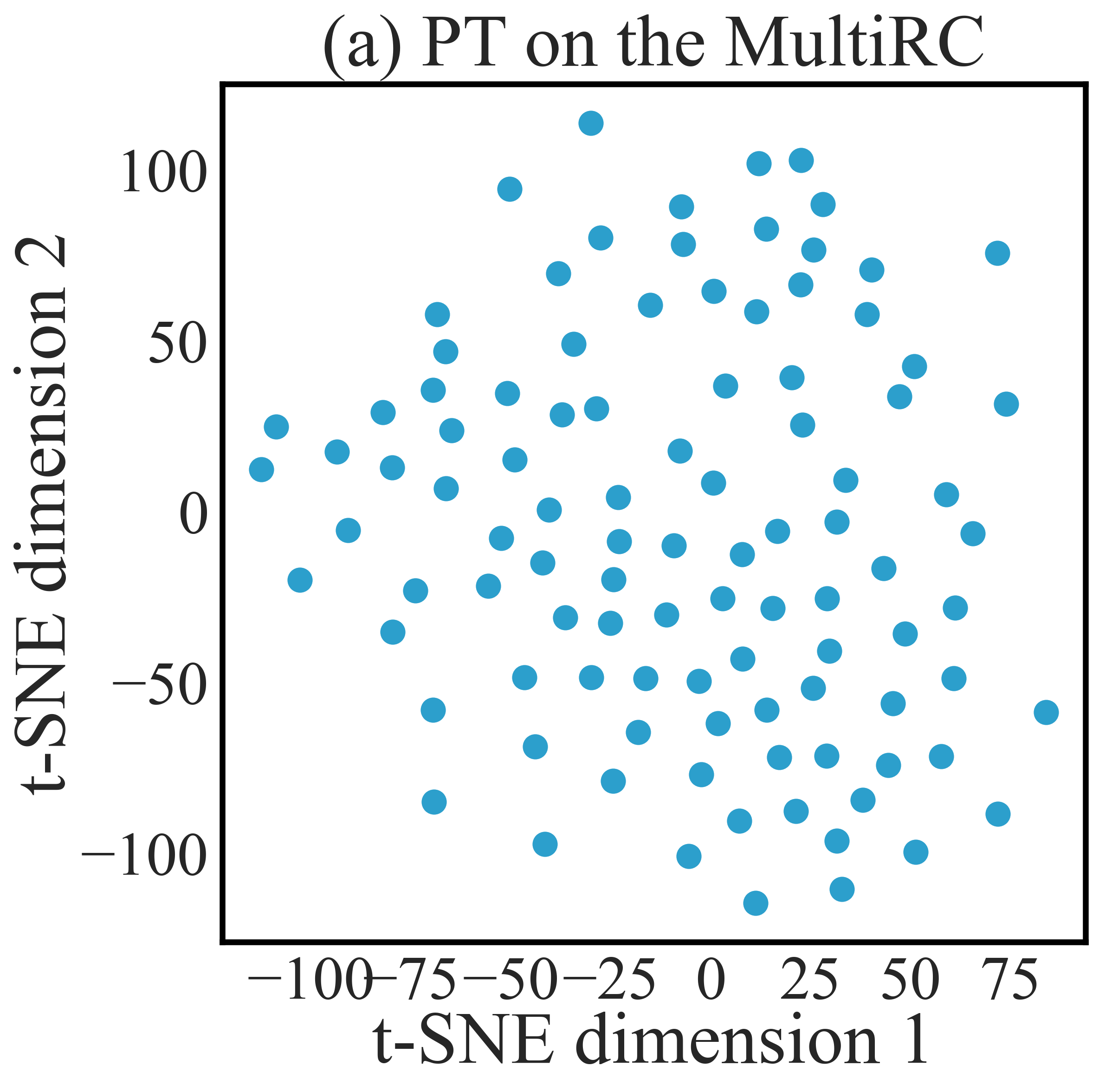}}
  \subfloat{
   \includegraphics[scale=0.188, trim={0mm 0mm 0mm 0mm}]{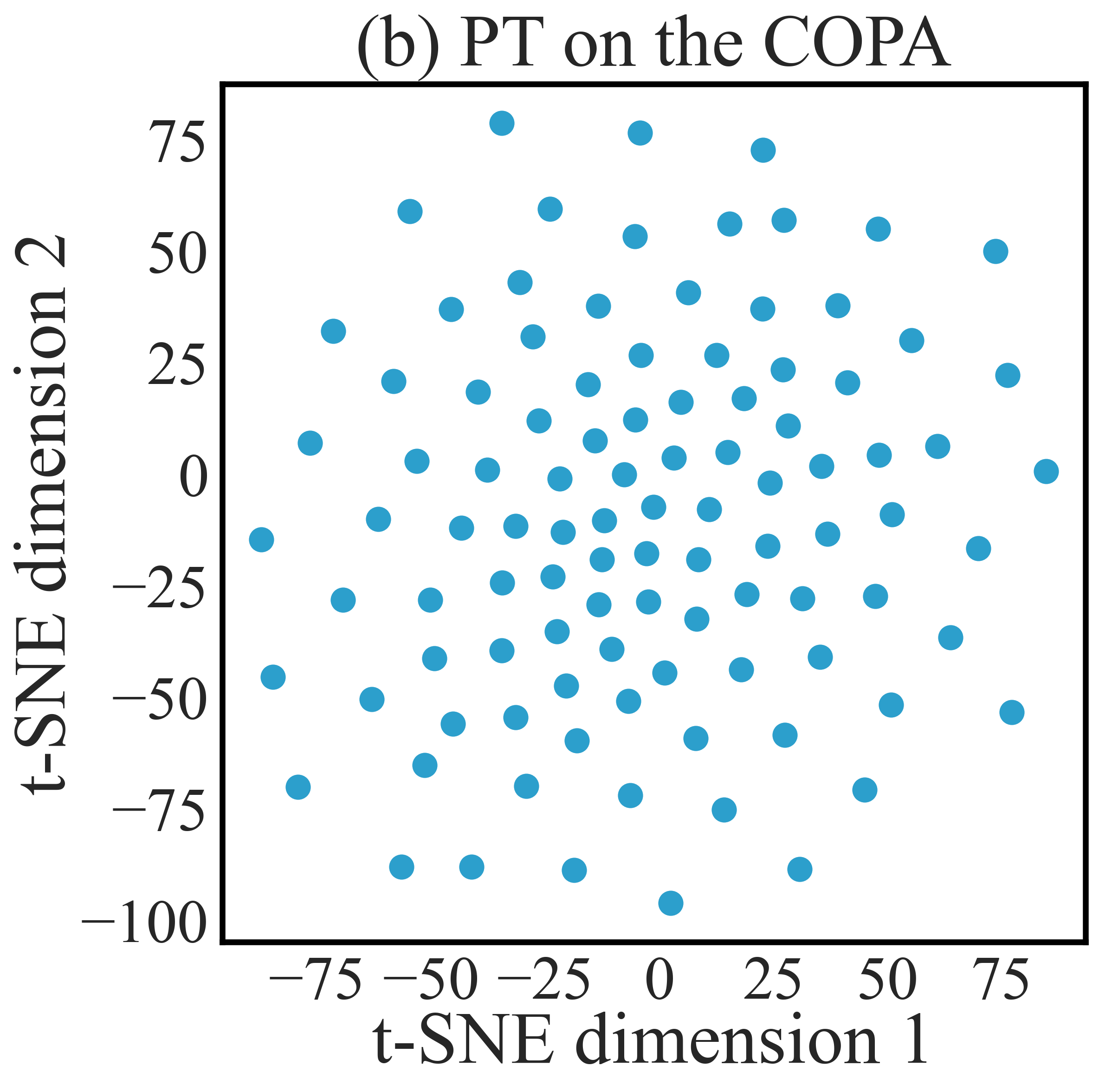}}
    \caption{(a) and (b) show the t-SNE clustering visualizations of the original prompt tuning prompts after training on the MultiRC and COPA datasets using the T5-Base model. Source prompt tokens are initialized from sampled vocabulary and length is set to 100. }
    \label{fig_tsne}
\end{figure}

\section{Introduction}
Pre-trained language models (PLMs) possess powerful learning capabilities to extract complex features and patterns from vast amounts of data \citep{devlin2019bert,radford2019language}. 
In recent years, as the scale of large PLMs has rapidly expanded, computational costs have surged dramatically \citep{lan2024efficient}. Although full fine-tuning parameters of PLMs yields satisfactory results, it has become impractical, e.g., PaLM has 540B parameters and requires 6144 TPU v4 chips to train for 1200 hours \citep{chowdhery2023palm}.

Parameter-Efficient Fine-Tuning (PEFT) methods attempt to bridge this gap by achieving performance comparable to full fine-tuning with minimal computational resources and time costs~\citep{houlsby2019parameter,hu2021lora,lester2021power}. Among these methods, prompt tuning (PT) stands out for its efficiency and flexibility. It freezes the model parameters and exclusively trains the soft prompt tokens attached to the model's input, delivering performance on par with full fine-tuning \citep{lester2021power,xiao2023decomposed,razdaibiedina-etal-2023-residual, lan2024efficient}. Notably, PT's trainable parameters are much lower than other PEFT methods (e.g., Adapter~\citep{houlsby2019parameter} and LoRA \citep{hu2021lora}) and do not grow exponentially as the scale of PLMs. 

\begin{figure}[t]
\centering
  \includegraphics[scale=0.18]{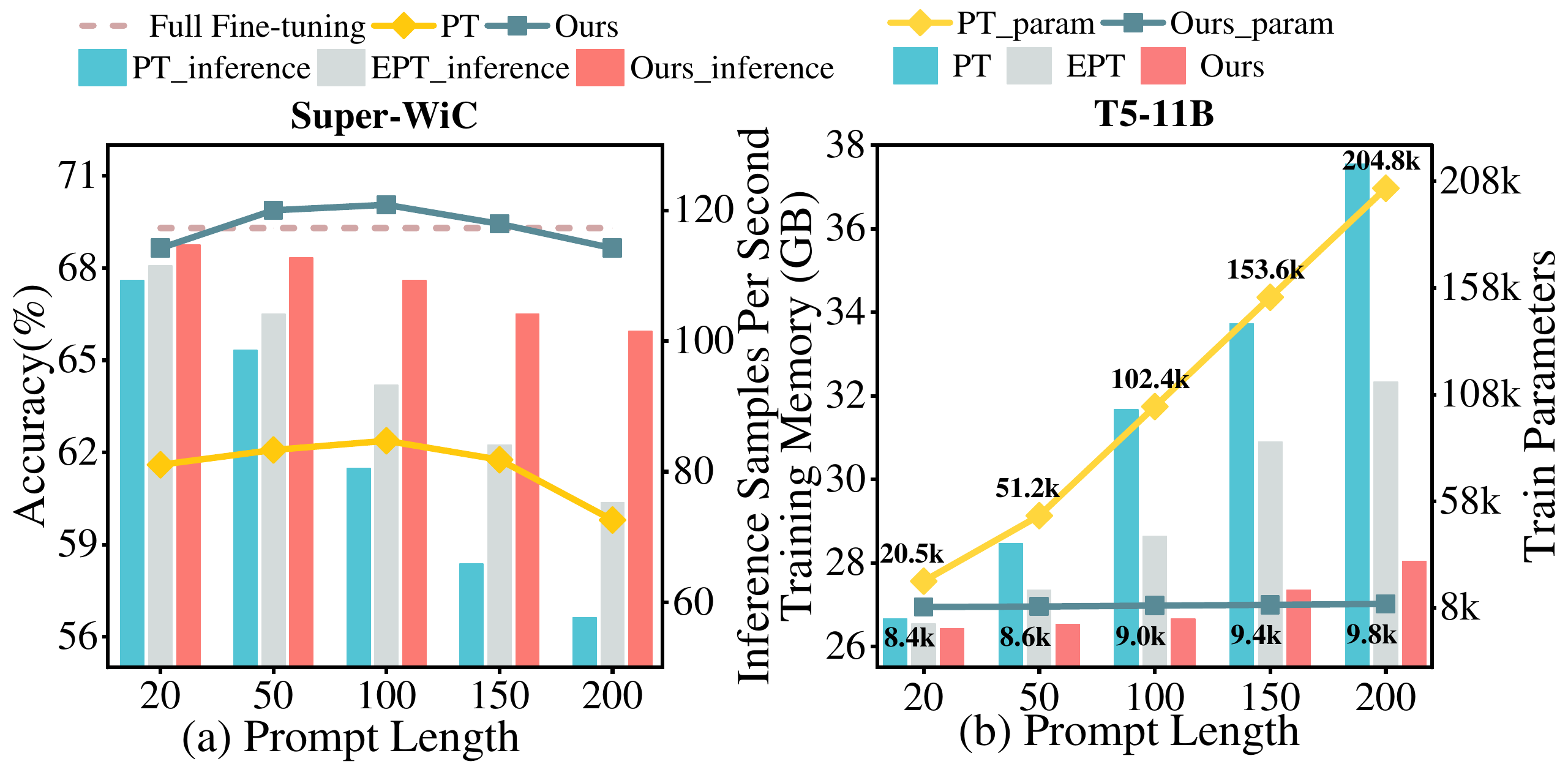}
    \caption{(a) Average performance on the T5 models across the SuperGLUE benchmark. (b) Impact of prompt length on performance and trainable parameters in the WiC dataset of the SuperGLUE benchmark.}
    \label{fig_total}
\end{figure}
 
Although these methods provide undeniable contributions, existing PT-based methods still suffer from two main challenges: \textbf{First}, in various tasks, PT initialization methods exhibit high discreteness, lacking exploration of the intrinsic semantic associations between tokens. The two mainstream prompt initialization methods—random initialization and initialization from sampled vocabulary (i.e., the 5,000 most common tokens)—help the model explore a broader parameter space during training. The sample vocabulary initialization, in particular, is widely utilized in pre-training for transfer learning \citep{vu2022spot,asai2022attempt} and multi-task learning \citep{wang2022multitask,zhong2024panda} due to its more informative nature. Nevertheless, as shown in Figure \ref{fig_tsne}, we found that although this informative initialization provides prompt tokens with rich knowledge, these tokens remain isolated without establishing semantic connections after training. In NLP tasks, intrinsic semantic associations form the context and meaning of language, helping models better understand and represent complex language structures \citep{mikolov2013efficient,devlin2019bert,Vaswani2017}. Prompt tokens leverage semantic knowledge to guide PLMs in producing outputs that better meet task requirements. Clearly, this intrinsic semantic association is essential for PT. This high discreteness overlooks capturing intrinsic semantic associations among soft prompt tokens, limiting the model's knowledge representation capabilities. \textbf{Second}, although PT does not require training the parameters of PLMs, adding soft prompts increases the total length of input embeddings. Figure \ref{fig_total} reveals the relationship between computational cost, trainable parameters, and prompt length. Previous research has shown that a long prompt 100 yields optimal PT performance, enabling PLMs to adapt to complex downstream tasks \citep{lester2021power,razdaibiedina-etal-2023-residual,xiao2023decomposed}. However, such a length renders PT inefficient. This inefficiency stems from the inherent high complexity of PLMs (e.g., the quadratic complexity of Transformers) \citep{Vaswani2017} and the fact that the storage of gradients and optimizer states is closely related to the number of trainable parameters \citep{guo2021parameter}. While some approaches \citep{xiao2023decomposed, shi2024dept,lan2024efficient} attempt to optimize standard PT, they still struggle with inefficiency and suboptimal performance.

To address the aforementioned challenges, we propose a novel efficient and effective low-parameters
prompt tuning (\textbf{LAMP}) method through prompt decomposition and compressed outer product.
Our motivation stems from the soft prompt exhibiting high dispersion and the ``intrinsic rank" \citep{aghajanyan2021intrinsic, hu2021lora} in PEFT, which indicates that model fine-tuning can occur in a low intrinsic dimensionality space. Specifically, LAMP first employs Truncated singular value decomposition (SVD) with its inherent structure—two low-dimensional singular vectors and singular values—to transform the loosely related semantic knowledge in PT tokens into a more structured and interrelated form, while simultaneously reducing trainable parameters. It then aggregates the semantic knowledge from the Truncated SVD and leverages the compressed outer product to enable multi-level interactions of intrinsic semantics, enhancing the model's knowledge representation. Finally, LAMP reduces computational load by applying average pooling, which does not increase training parameters. 
Figure \ref{fig_total} demonstrates that the longer the prompt length, the more significant the reduction in computational cost and memory usage achieved by LAMP.

The main contributions of this paper are:
\begin{itemize}[leftmargin=*]
\item Our empirical study reveals that tokens in soft prompts exhibit high dispersion during training, lacking inherent semantic interactions among tokens to assist the model in comprehending and handling complex tasks, thereby limiting the model's knowledge representation capability.
\item We propose a novel low-parameter prompt tuning (abbreviated as LAMP) method that captures potential semantic interactions between prompt tokens through prompt decomposition and compressed outer product. LAMP achieves robust performance while significantly reducing computational costs (e.g., training time, memory usage, and trainable parameters).
\item We comprehensively evaluated LAMP on the SuperGLUE benchmark. Experimental results demonstrate that LAMP outperforms other state-of-the-art PT methods and remains effective in few-shot scenarios. Notably, on the T5-11B model, LAMP improved performance by 5.59\% compared to the vanilla PT while also increasing inference speed by 31\%, reducing trainable parameters by 91.21\%, shortening training time by 23.64\%, and lowering memory usage by 24.49\%.
\end{itemize}

\section{Method}

\begin{figure*}[h]
\centering
  \includegraphics[scale=0.85]{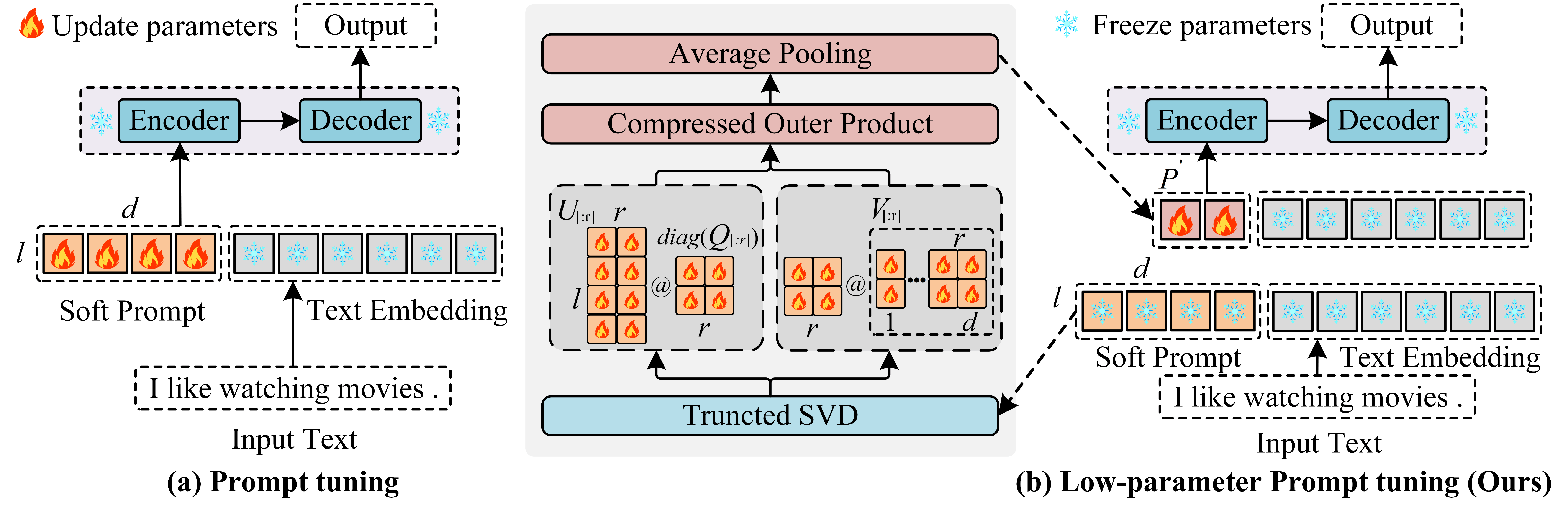}
    \caption{(a) Conventional prompt tuning \citep{lester2021power}. (b) The overview of our proposed LAMP. It decomposes the vanilla prompt to construct a new low-dimensional prompt, captures the intrinsic semantic associations between prompt tokens, and finally reduces computational costs through average pooling.}
    \label{fig_LAMP}
\end{figure*}

\subsection{Preliminaries}
\label{PT}

\ \\ \noindent
\textbf{Prompt Tuning.} PT can maintain parameter efficiency as model size scales. This approach ensures that trainable parameters does not increase dramatically with model expansion, making it a preferred choice for many applications \citep{shi2024dept, lan2024efficient}. Let labelled training data $(\boldsymbol{X}, \boldsymbol{Y})=\left\{\boldsymbol{x}_i, \boldsymbol{y}_i\right\}_{i=1}^N$ for one target task $\mathcal{T}$, the number of training data is $N$. The total parameters of PLM is $\Theta$ and each input text is $\boldsymbol{x}_i$. The embedding of $\boldsymbol{x}_i$  is represented as $\mathbf{E}_{i}\in \mathbb{R}^{{m \times d}}$, where $m$ is maximum sequence length and $d$ is the dimension of input embedding. The target prompt $\mathbf{P} \in \mathbb{R}^{{l \times d}}$ is initialized, with $l$ as the hyper-parameter determining the length of the soft prompt. This prompt is then concatenated with the fixed embedding $\mathbf{E}_{i}\in \mathbb{R}^{{m \times d}}$. $\mathbf{E}_{i}$ remains unchanged during gradient updates in training, resulting in a new input embedding $\left [ \mathbf{P};\mathbf{E}_{i} \right]\in \mathbb{R}^{(l+m)\times d}$. The formulation for the target task is as follows:
\begin{equation}
    \mathcal{L}_{p} = -\sum_{i}\log P \left(\mathbf{y}_{i}|\left[\mathbf{P};\mathbf{E}_{i} \right];\Theta \right)
\label{eq2}
\end{equation}
where $\mathcal{L}_{p}$ is a loss function only optimized with the prompt $\mathbf{P}$. $P(\cdot)$ is maximizing the conditional probability of PT. The overall structure of PT is shown in Figure \ref{fig_LAMP}(a).

\subsection{LAMP: Low-parameters Prompt Tuning}
\textbf{Prompt initialization.} We initialize the source soft prompt $\mathbf{P}$ from sampled vocabulary \citep{lester2021power,vu2022spot,razdaibiedina-etal-2023-residual, asai2022attempt}, a strategy that embeds more semantic richness and prior knowledge than mere random initialization. Inadequate initialization can result in the discovery of suboptimal local minima, thus impairing the model's generalization capabilities.
\\ \ \noindent
\textbf{Prompt decomposition.} Our innovative motivation stems from two aspects: 
(1) Soft prompt tokens exhibit high dispersion (as shown in Figure~\ref{fig_tsne}), neglecting the knowledge interactions between tokens;
(2) Soft prompt also exhibit low "intrinsic rank" behavior \citep{xiao2023decomposed}.
Inspired by these findings and the core idea of Truncated SVD \citep{hansen1987Truncated}, we employ Truncated SVD to reduce training parameters by decomposing the soft prompt $\mathbf{P}\in \mathbb{R}^{l\times d}$. The original SVD of $\mathbf{P}$ is formulated as follows:
\begin{equation}
   \mathbf{P}=\mathbf{U}\text{diag}(\mathbf{Q})\mathbf{V}^\top
\end{equation}
where $\mathbf{U} \in \mathbb{R}^{l\times min(l,d)}$, $\mathbf{V}\in \mathbb{R}^{d\times min(l,d)}$ are the singular vectors with orthogonal columns, $\mathbf{U}$ and $\mathbf{V}$ transforms highly dispersed tokens from the original PT into interrelated representations. $\mathbf{\mathbf{Q}} \in \mathbb{R}^{min(l,d)}$ comprises the singular values arranged in descending order (the larger the singular value, the more information it contains). The operation $\text{diag}(\mathbf{Q})$ converts $\mathbf{Q}$ into a diagonal matrix, and $V^\top$ represents the transpose of $\mathbf{V}$.
\\ \ \noindent
\textbf{Prompt reconstruction.} We define the soft prompt's low "intrinsic rank" as $r$ and select the top-$r$ singular values, $\mathbf{Q}_{[:r]}$, which contain a rich amount of information arranged in descending order. The remaining singular values, $\mathbf{Q}_{[r:]}$, are discarded and do not participate in training/inference. Consequently, as the $\mathbf{Q}$ dimension changes, singular values and vectors are redefined as $\left \{ \mathbf{U}_{[:r]}\in \mathbb{R}^{l\times r},\mathbf{Q}_{[:r]}\in \mathbb{R}^{r},\mathbf{V}_{[:r]}\in \mathbb{R}^{d\times r}\right \}$. We can approximate the original information by storing only $r$ (where $r \ll d$) singular values and their corresponding low-parameters singular vectors, achieving parameter compression, which is also why LAMP adopts Truncated SVD.

The trainable parameters are now reduced from `$l \times d$' to `$l \times r + r + r \times d$'. In experiments, we provide a detailed explanation of hyperparameter $r$ and its impact on model performance. 
For instance, on the Llama2-7B \citep{touvron2023llama}, when the prompt length increases from 100 to 500, traditional PT requires training 2,048k parameters, whereas our method, LAMP, only requires training ($500 \times 8+8+8 \times 4096=36.8$k) parameters. LAMP's advantage becomes more pronounced with longer prompt lengths or larger model scales, as it significantly reduces the computational cost by decreasing trainable parameters. LAMP has established a solid foundation for PT-based methods to excel across various domains.
\\ \ \noindent
\textbf{Compressed outer product.} Although using Truncated SVD to reduce the trainable parameters in prompt tuning is promising, directly applying dot products on the decomposed singular values and vectors can only partially capture the intrinsic associations between tokens. Specifically, dot products' inherent linear nature limits their ability to fully express the more complex, nonlinear interactions among prompt tokens. Considering that outer products can mine richer and more complex high-order interactions, we utilize the compressed outer product to further explore the intrinsic semantic associations between tokens in prompt tuning. 

Firstly, we utilize the dot product of singular values and singular vectors as the initialization input for the compressed outer product module:
\begin{equation}
   \mathbf{M}=\mathbf{U}_{[: r]} \operatorname{diag}\left(\mathbf{Q}_{[: r]}\right) \in \mathbb{R}^{l \times r},
\end{equation}
\begin{equation}
   \mathbf{I}=\operatorname{diag}\left(\mathbf{Q}_{[: r]}\right) \mathbf{V}_{[: r]}^\top \in \mathbb{R}^{r \times d} .
\end{equation}
This approach enables an initial aggregation of knowledge features between tokens, which helps to effectively represent and explore the underlying semantic knowledge associations. Due to compressed outer product can maintains the high-order structure while facilitating multiple layers of intrinsic semantic interactions. This approach effectively restores and enhances the complex information structures that might be lost in Truncated SVD, thereby enriching the knowledge representation capabilities of prompt tokens. The compressed outer product is formulated as follows:
\begin{equation}
\mathbf{C} = \sum_{i=1}^r \mathbf{M}_{[:, i]} \otimes \mathbf{I}_{[i, :]} \in \mathbb{R}^{l \times d}
\end{equation}
where $\mathbf{M}_{[:, i]}$ is the $i$-th column vector of $\mathbf{M}$, $\mathbf{I}_{[i, :]}$ is the $i$-th row vector of $\mathbf{I}$, $\otimes$ denotes the outer product of two vectors. $\mathbf{C} \in \mathbb{R}^{l \times d}$ is the resultant prompt after summing all the outer products. The introduction of compressed outer products enhances the representational power of prompt tuning. This approach enables the soft prompt to more effectively adapt to different downstream tasks through deep interactions between prompt tokens.

While introducing compressed outer product does not create new trainable parameters, it does entail a slight increase in computational overhead due to its engagement in more complex higher-order interactions. Additionally, given the transformers' quadratic complexity, the prompt's length is proportional to the training duration. We consider employing average pooling operation to reduce training time: 
\begin{equation}
    \mathbf{P'}_{i,j} = \frac{1}{p} \sum_{k=0}^{p-1} \mathbf{C}_{i\ast p+k, j}
\end{equation}
$\mathbf{P'}_{i,j}$ represents the elements of the tensor $\mathbf{P'}\in \mathbb{R}^{l/p \times d}$ after averaging pooling. This operation effectively compresses the $l$ elements along the first dimension into $l / p$. 
It is also noteworthy that we explored a self-attention pooling strategy to dynamically filter prompt tokens in Appendix \ref{app_Self-Attention Pooling}; however, this strategy was not very effective and introduced additional trainable parameters.

\subsection{Training and Inference}
Only the parameters of $\small \mathbf{U}_{[:r]}\in \mathbb{R}^{l\times r}$, $\small \mathbf{Q}_{[:r]}\in \mathbb{R}^{r}$, and $\small \mathbf{V}_{[:r]}\in \mathbb{R}^{d\times r}$ are optimized during the training process, while the backbone model (i.e., $\Theta$ and $\mathbf{E}_{i}$) remained frozen as Figure~\ref{fig_LAMP}(b). The reconstructed prompt $\mathbf{P'}$ is inserted before the input text embeddings. By $\mathbf{P'}$, Eq.\ref{eq2} is displaced by:
\begin{equation}
\mathcal{L}_{PT} = -\sum_{i}\; \log P(\mathbf{y}_{i}|[\mathbf{P'};\mathbf{E}_{i}];\Theta)
\label{eq5}
\end{equation}
where $[\mathbf{P'};\mathbf{E}_{i}] \in \mathbb{R}^{(l/p+m)\times d}$ is a input embedding of PLMs through the connection of $\mathbf{P'}$ and $\mathbf{E}_{i}$.

\section{Experiments}
In this section, we will answer these key research questions by conduct extensive experiments: 
\textbf{RQ1:} How does our LAMP performance compare with other SOTA baselines across different model scales and datasets?
\textbf{RQ2:} How do few-shot adaptability and hyper-parameters optimization influence the LAMP?
\textbf{RQ3:} How will the feature space of the soft prompt change after considering the intrinsic semantic associations between tokens?

\subsection{Evaluation Datasets and Metrics}
\textbf{Evaluation Datasets:} Building upon prior studies in prompt tuning \citep{xiao2023decomposed}, we employ eight NLP tasks from the SuperGLUE \citep{wang2019superglue} and GLUE \citep{wang2018glue} benchmark and conduct multi-aspect experiments to evaluate the high efficiency and effectiveness of LAMP. The SuperGLUE benchmark includes more complex and challenging tasks among eight datasets than GLUE \citep{wang2018glue}: CB ~\citep{de2019commitmentbank}, WSC ~\citep{levesque2012winograd}, COPA \citep{roemmele2011choice}, RTE \citep{giampiccolo2007third}, WiC ~\citep{pilehvar2019wic}, BoolQ ~\citep{clark2019boolq}, MultiRC ~\citep{khashabi2018looking} and ReCoRD \citep{zhang2018record}. MNLI \citep{williams2018broad}, QNLI \citep{rajpurkar2016squad}, SST-2 \citep{socher2013recursive} and  MRPC\citep{bill2005automatical} in GLUE benchmark. More details about datasets in Appendix \ref{app_Dataset Details}.
\ \\ \noindent
\textbf{Metrics:} Consistent with previous work \citep{razdaibiedina-etal-2023-residual,xiao2023decomposed}, the evaluation metric is F1 for MultiRC and ReCoRD, the evaluation metric is Accuracy for other tasks. 

\subsection{Baselines and Models}
We compare LAMP with the following baseline approaches: \textbf{Full Fine-tuning}, which updates all parameters of PLMs; \textbf{PT-based methods}, where PT \citep{lester2021power} inserts trainable continuous vectors, known as soft prompt, before the model's input, and its most advanced variants include Residual PT \citep{razdaibiedina-etal-2023-residual}, DePT \citep{shi2024dept}, EPT \citep{lan2024efficient} and DPT \citep{xiao2023decomposed}; \textbf{LoRA-based methods} include PiSSA \citep{meng2024pissa}, rsLoRA \citep{kalajdzievski2023rsLoRA}, LoRA+ \citep{hayoulora+}, DoRA \citep{liudora} and LoRA-GA \citep{wang2024loraGA} . More details about Baselines can be found in Appendix \ref{app_Baselines Details}.

We aim to explore a high-performance PEFT method that minimizes trainable parameters. \textbf{Trainable parameters are a crucial factor in our selection of baselines}; hence, methods with more significant trainable parameters and modifying the transformer layers are not included as baselines for comparison. Such as Adapter \citep{houlsby2019parameter} (prompt length is 100, 76.8k vs. 1.9M for T5-base) and its variant methods. Notably, EPT \citep{lan2024efficient} has demonstrated superior performance compared to these PEFT methods. Furthermore, Xprompt \citep{ma2022xprompt} underwent rewinding training, and transfer learning \citep{vu2022spot,asai2022attempt} and multi-task learning \citep{wang2022multitask} require pre-training. These methods are not directly comparable to LAMP.
\subsubsection{Models Size}
PT tends to underperform in smaller-scale models. Thus, we conducted primary experiments employing three T5 model variants \citep{raffel2020exploring} (Small, 60M; Base, 220M; and Large, 770M) and validated the effectiveness of LAMP using T5-11B and Llama2-7B \citep{touvron2023llama}. 

\subsection{Training Details}
T5 model as the backbone for our experiments; the hidden dimensions for the T5-base, T5-small, and T5 large are 512, 768, and 1,024, respectively. Following the experimental setup from \citet{xiao2023decomposed}, we set the soft prompt length to 100, rank $r=8$ in Truncated SVD and batch size is 16. 
The models are trained 100 epochs using the AdamW \citep{loshchilov2018decoupled} optimizer with an initial learning rate of 0.3. We employ the double quantization operation in QLoRA\citep{dettmers2023qlora} for T5-11B and Llama2-7B.
Other training details in Appendix \ref{app_Implementation_Details}.

\begin{table*}[t]
  \centering
\renewcommand{\arraystretch}{0.95}
\setlength{\tabcolsep}{0.1cm}
\begin{tabular}{l|l|cccccccc|>{\columncolor{gray!25}}c}
    \toprule
    \multirow{2}[2]{*}{\textbf{Method}} & \multirow{2}[2]{*}{\textbf{Params.}} & \textbf{CB} & \textbf{WSC} & \textbf{COPA} & \textbf{RTE} & \textbf{WiC} & \textbf{BoolQ} & \textbf{MultiRC} & \textbf{ReCoRD} & \multicolumn{1}{c}{{\textbf{Average}}} \\
          &  & Acc.   & Acc.   & Acc.   & Acc.  & Acc.  & Acc.  & F1 & F1 & (\%)  \\
    \midrule
    \multicolumn{11}{c}{\textbf{T5-Small}} \\
    Fine-Tuning$^{\dagger}$ & 60M   & 89.28  & 67.94  & 59.00  & 72.56  & 68.18  & 77.06  & 66.98  & 55.64  & 69.58  \\
    \midrule
    Prompt Tuning & 51K   & 71.43  & 59.62  & 58.33  & 66.91  & 63.95  & 66.12  & 63.31  & 50.11  & 62.47  \\
    Residual PT$^{\dagger}$ & 462K  & 72.02  & 63.14  & 56.66  & 67.02  & 60.96  & \textbf{73.35 } & \underline{65.12}  & 53.08  & 63.91  \\
    DePT  & 51K   & 75.00  & \textbf{67.31}  & 52.66  & 65.22  & 62.70  & 66.00  & 61.94  & \textbf{56.71 } & 63.44  \\
    EPT   & 51K   & 78.57  & \textbf{67.31}  & 55.66  & \textbf{71.01}  & \textbf{67.39} & 69.17  & 64.46  & \underline{54.14}  & \underline{65.96}  \\
    DPT   & 6K    & \underline{78.85}  & 60.53  & \underline{59.33}  & \underline{70.40}  & 64.26  & 72.17  & 64.61  & 53.60  & 65.47 \\
    \midrule
    \textbf{LAMP(ours)} & 5K   & \textbf{83.93} & \textbf{67.31}  & \textbf{60.66}  & 69.31 & \underline{66.46} & \underline{72.97} & \textbf{66.25} & 53.80 & \textbf{67.59} \\
    \midrule
    \multicolumn{11}{c}{\textbf{T5-Base}}\\
    Fine-Tuning$^{\ddagger}$ & 220M  & 91.70  & 81.70  & 60.00  & 84.50  & 69.30  & 82.30  & 76.90  & 80.90  & 78.41  \\
     \midrule
    Prompt Tuning & 77K   & 78.57  & 61.54  & 55.00  & 67.63  & 62.38  & 77.00  & 72.37  & 71.32  & 68.27  \\
    Residual PT$^{\dagger}$ & 693K  & 77.37  & \underline{67.94}  & \underline{56.66}  & \underline{81.70}  & 66.87  & 80.00  & 72.11  & 72.21  & 71.86  \\
    DePT  & 77K   & 82.14  & 67.31  & 54.33  & 73.91  & 65.20  & 79.02  & \textbf{72.70} & 70.80  & 70.68  \\
    EPT   & 77K   & \underline{85.71}  & 67.31  & \underline{56.00}  & 78.99  & 67.71  & 79.14  & \underline{72.62}  & 71.15  & \underline{72.33}  \\
    DPT   & 9K    & 78.56  & 67.30  & \underline{56.66}  & 79.42  & \underline{68.49}  & \textbf{80.28}  & 72.50  & \textbf{72.56}  & 71.97  \\
    \midrule
    \textbf{LAMP(ours)} & 7K   & \textbf{94.64} & \textbf{68.42}  & \textbf{58.66}  & \textbf{83.39} & \textbf{70.06} & \underline{80.24} & 72.72 & \underline{72.55} & \textbf{75.09} \\
    \midrule
    \multicolumn{11}{c}{\textbf{T5-Large}}\\
    Fine-Tuning$^{\ddagger}$ & 770M  & 94.30  & 88.50  & 87.00  & 90.60  & 73.50  & 88.30  & 85.40  & 89.20  & 87.10  \\
     \midrule
    Prompt Tuning & 102K  & 82.35  & 65.38  & \underline{57.33}  & 88.45  & 70.69  & 84.28  & 76.37  & 74.36  & 74.90  \\
    Residual PT$^{\dagger}$ & 925K  & 73.21  & \underline{70.50}  & \textbf{62.66} & \underline{88.92}  & \underline{72.25}  & \underline{85.04}  & 76.46  & \underline{84.36}  & 76.67  \\
    DePT  & 102K  & 85.71  & 67.31  & 50.66  & 83.33  & 68.97  & 83.24  & 75.76  & 74.03  & 73.63  \\
    EPT   & 102K  & \underline{89.29}  & 68.30  & 54.00  & 86.33  & 71.79  & 84.77  & 76.62  & 73.94  & 75.63  \\
    DPT   & 11K   & \underline{89.29}  & 65.79  & \textbf{62.66} & 88.45  & 71.63  & 84.53  & \underline{76.72}  & 84.35  & \underline{77.93}  \\
    \midrule
    \textbf{LAMP(ours)} & 9K   & \textbf{98.21} & \textbf{78.95}& \underline{57.33}  & \textbf{90.61} & \textbf{73.35}& \textbf{85.11} & \textbf{76.94} & \textbf{84.56}& \textbf{80.63} \\
    \bottomrule
    \end{tabular}%
\caption{For the performance comparison on the SuperGLUE benchmark, all experimental results are based on the T5-Small, T5-Base, and T5-Large. All scores represent the mean across three runs using distinct random seeds. 
$^{\dagger}$ sourced from \citet{xiao2023decomposed}. $^{\ddagger}$ sourced from \citet{aribandi2021ext5}. The best result is marked in bold. The second-highest result is indicated by an underline. }
  \label{tab1}%
\end{table*}%

\begin{table}[t]
\renewcommand{\arraystretch}{1}
\setlength{\tabcolsep}{0.3cm}
  \centering
     \begin{tabular}{c|ccc}
    \toprule
    \multicolumn{4}{c}{\textbf{K-Shot}} \\
    \midrule
    \textbf{Model} & \textbf{8} & \textbf{16} & \textbf{32} \\
    \midrule
    Prompt Tuning & 48.23 & 49.83 & 50.85 \\
    Residual PT & 52.95 & 57.57 & 58.50 \\
    DePT  & 49.83 & 50.31 & 49.93 \\
    EPT   & 50.42 & 50.71 & 53.51 \\
    DPT   & 56.26 & 55.60 & 57.72 \\
    \midrule
    \textbf{LAMP (ours)} & \textbf{57.21} & \textbf{58.14} & \textbf{59.25} \\
    \bottomrule
    \end{tabular}%
  \caption{Few-shot adaptation results with $k = \{8, 16, 32\}$ on SuperGLUE benchmark.}
  \label{few-shot}%
\end{table}%

\begin{figure*}[t]
\centering
  \includegraphics[scale=0.215]{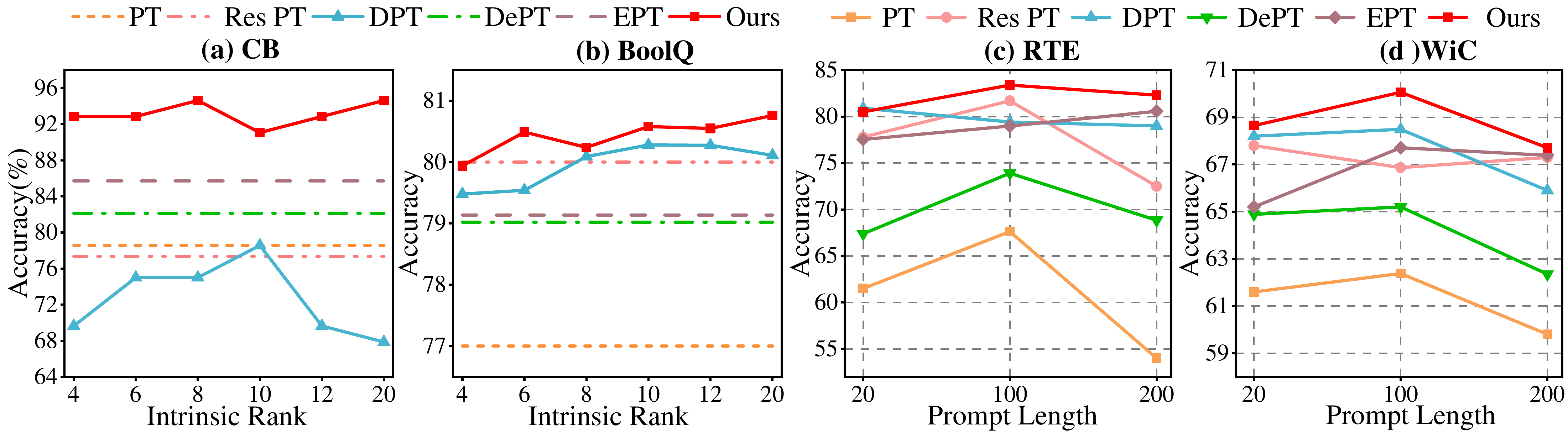}
    \caption{(a) and (b), the performance of all baselines with the number of inherent ranks $r \in\{4, 6, 8, 10, 12, 20\}$ on the SuperGLUE benchmark. (c) and (d), the performance of different baselines varies with the prompt length $l \in \{20, 100, 200\}$. All results represent the average of three runs conducted with a different random seed.}
    \label{fig_rank_length}
\end{figure*}

\begin{figure}[t]
\centering
  \includegraphics[scale=0.147]{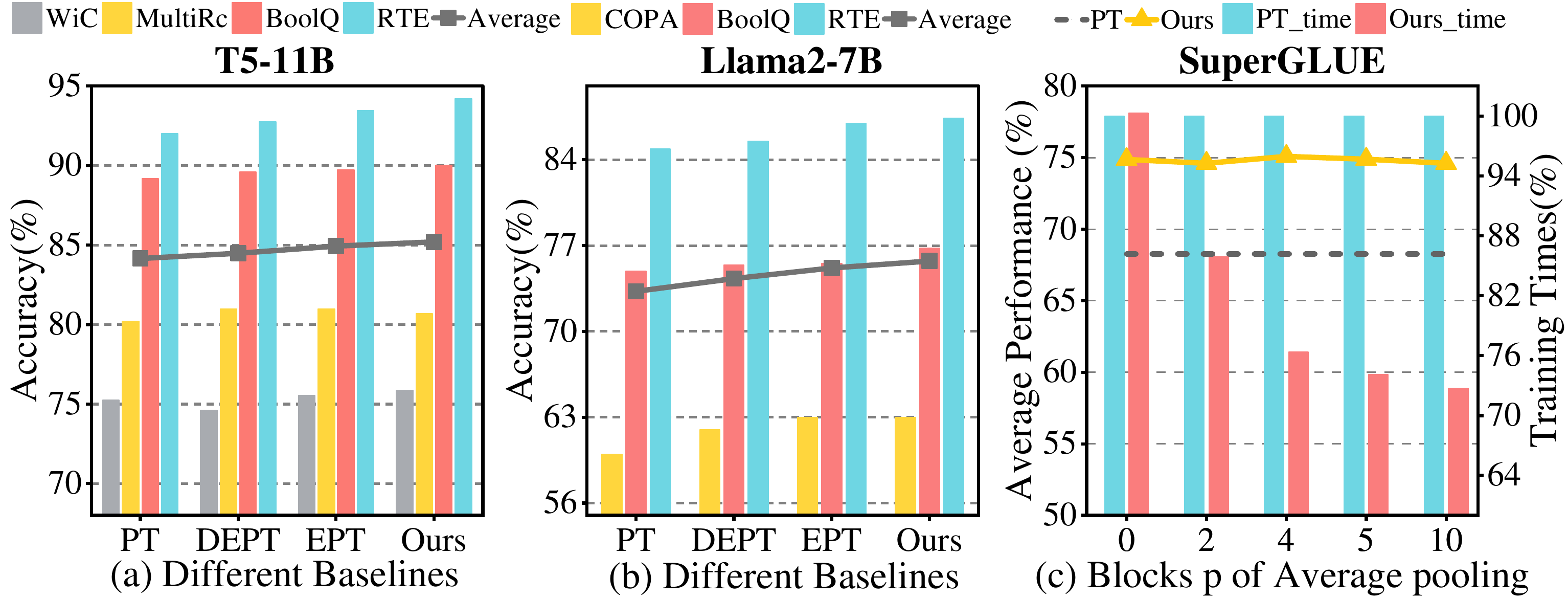}
    \caption{(a) and (b), the performance changes of different methods at various datasets on the T5-11B and Llama2-7B. (c), the variation of training time and performance in different average pooling blocks $p$.}
    \label{Fig_11b_llama}
\end{figure}

\subsection{Overall Performance Comparison (RQ1)}
Table \ref{tab1} presents a comparison of LAMP against baseline methods on the SuperGLUE benchmark using various T5 model sizes. Notably, LAMP requires the fewest training parameters and demonstrates exceptional average performance across different scales of T5 models. 
LAMP outperforms the original PT by 7.58\%, 9.08\%, and 7.11\% on T5-Small, T5-Base, and T5-Large, respectively. Detailed information on the standard deviation of LAMP can be found in Appendix \ref{app_Standard Deviation}. LAMP's performance improvement is attributed to enhancing the model's knowledge representation by uncovering the intrinsic semantic interactions between soft prompt tokens. LAMP achieves outstanding performance while significantly reducing training parameters and computational costs.

From the perspective of trainable parameters, LAMP and DPT are more efficient than other baselines; even though EPT outperforms DPT, EPT requires more trainable parameters. LAMP clearly outperforms DPT in several ways. First, DPT relies on randomly generated initial prompts that lack semantic richness, whereas LAMP leverages sample vocabularies to better aid the model in understanding complex language structures. Additionally, while DPT reduces trainable parameters, its prompt length remains at 100 when input into model, leading to inefficiency. In contrast, LAMP employs average pooling operations that neither harm performance nor increase trainable parameters, making it an efficient and effective novel prompt tuning method. \textbf{LAMP demonstrates superior performance and requires significantly fewer training parameters than the latest LoRA-based PEFT methods,} with detailed results provided in Appendix \ref{app_lora}.

\subsection{Ablation Experiment Analysis (RQ2)}
\label{subsec:few-shot}
\textbf{Few-shot adaptation.} 
Following the few-shot experimental setup of \citet{xiao2023decomposed}, we randomly sampled 8, 16, and 32 training examples. Table \ref{few-shot} presents the results of all baselines on the SuperGLUE benchmark. All results are averaged over three runs with different random seeds on the T5-base model. We found that LAMP outperforms other baselines on most datasets in the few-shot setting, demonstrating its effectiveness. The few-shot performance of various methods across different datasets is detailed in Appendix \ref{app_Few-shot Details}.
\ \\ \noindent
\textbf{Sensitivity of Rank Size.}
The intrinsic rank $r$ is the primary factor influencing the total number of trainable parameters in LAMP. We analyzed the impact of $r \in\{4, 6, 8, 10, 12, 20\}$ on LAMP performance using the T5-base model on the CB and BoolQ datasets within the SuperGLUE benchmark. As shown in Figure \ref{fig_rank_length}(a) and Figure \ref{fig_rank_length}(b), despite the minimal differences in trainable parameters of LAMP and DPT across different $r$ values, LAMP consistently outperforms DPT and other baseline methods in most scenarios. This demonstrates the effectiveness of incorporating semantic knowledge into LAMP. The details of how the intrinsic rank $r$ affects the changes in training parameters can be found in the Appendix \ref{app_Intrinsic Rank}.
\begin{figure*}[t]
\centering
    \subfloat{
		\includegraphics[scale=0.2]{Fig/a_PT_tsne_multirc.png}}
     \subfloat{
		\includegraphics[scale=0.2]{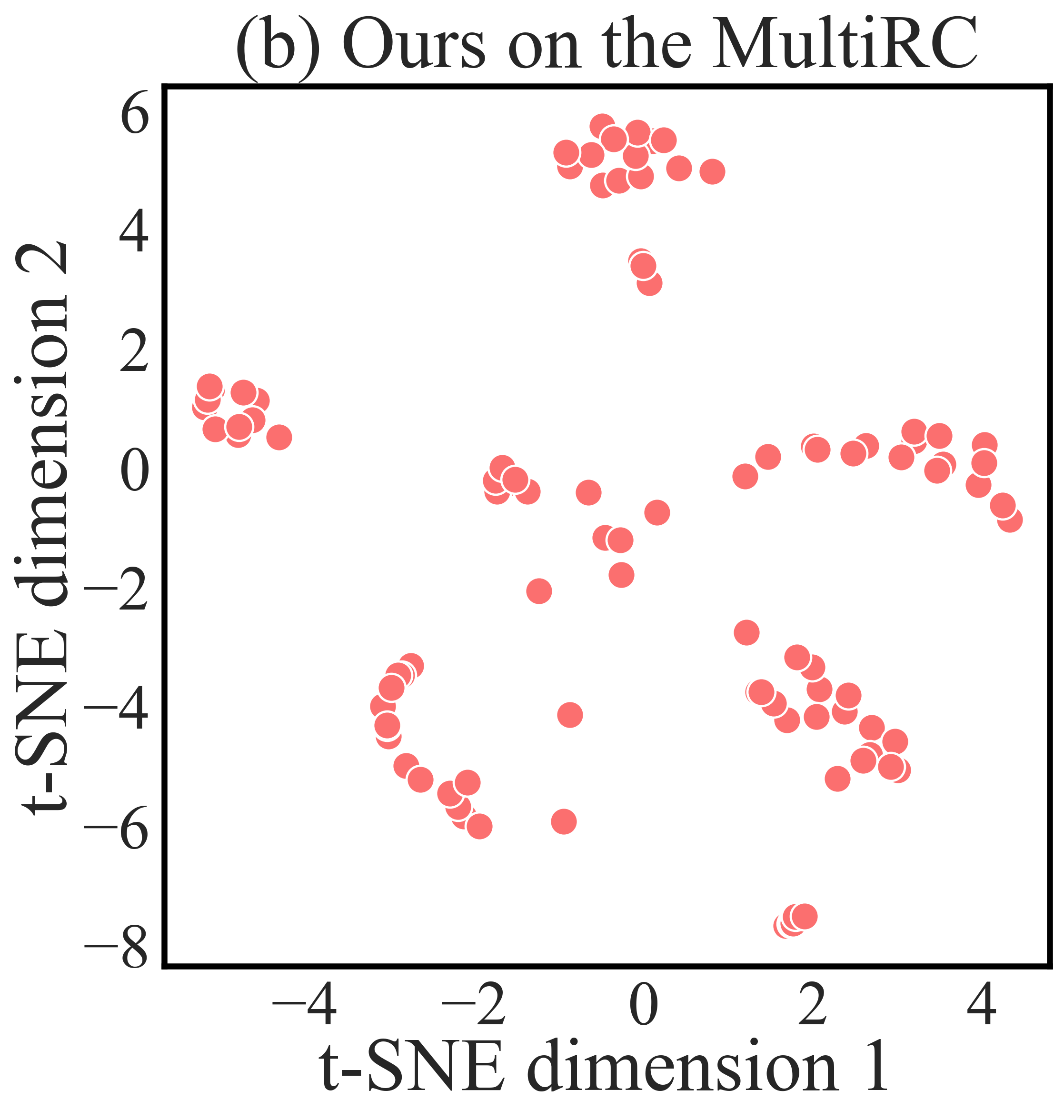}}
    \subfloat{
		\includegraphics[scale=0.2]{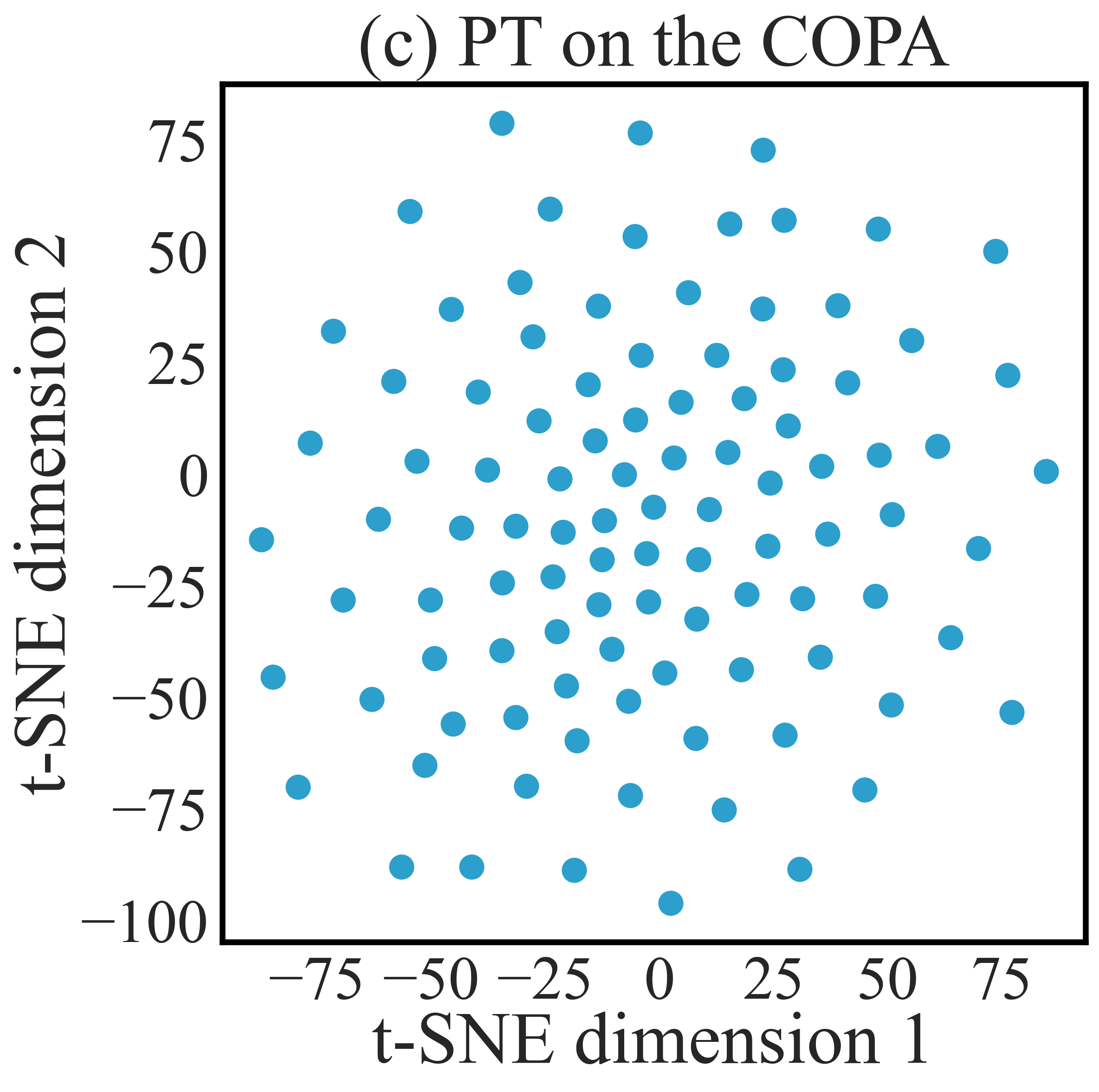}}
  \subfloat{
		\includegraphics[scale=0.2]{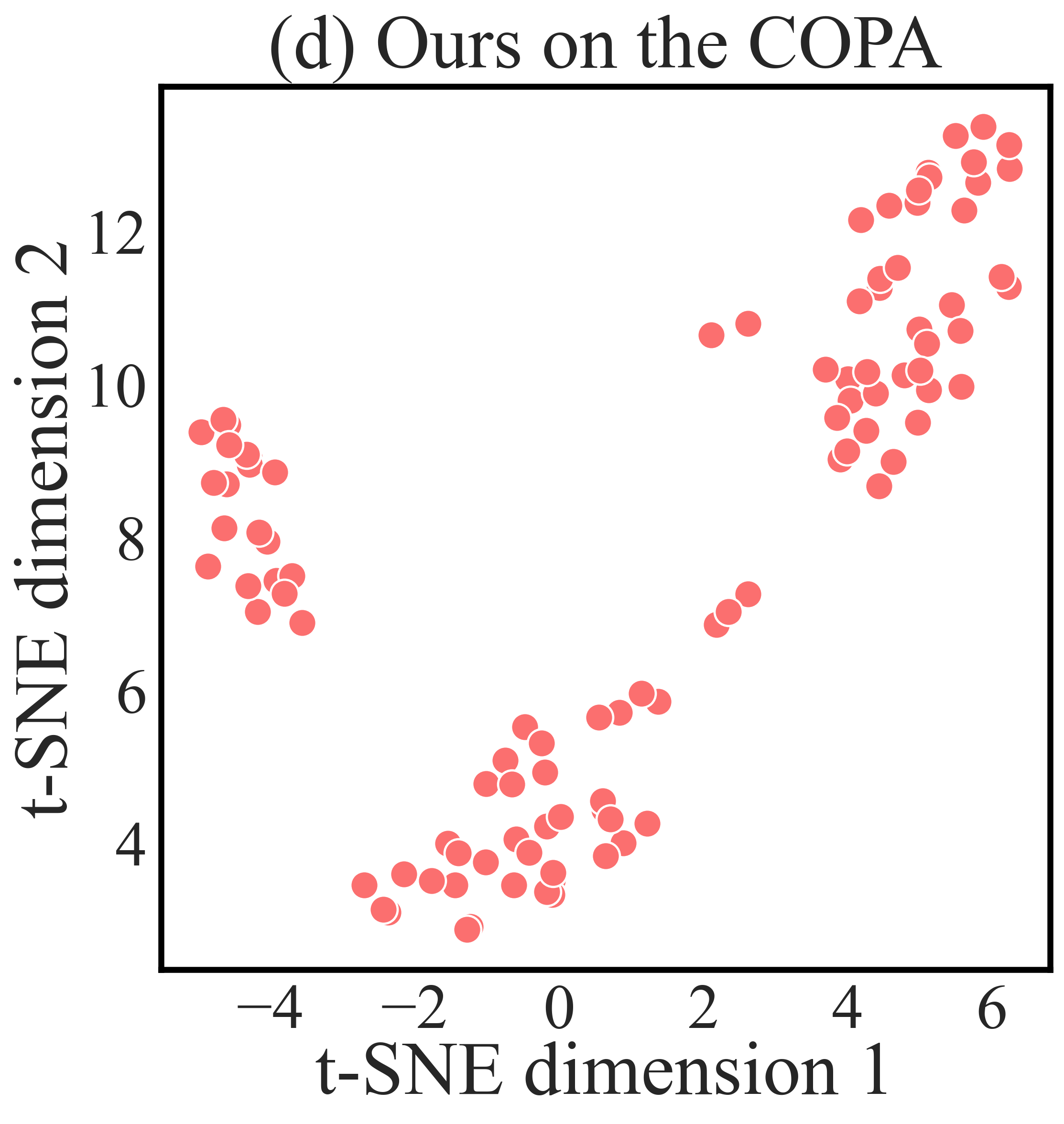}}
\caption{The comparison of the dispersion of PT tokens on the MultiRC and COPA datasets before and after considering the intrinsic semantic associations between soft prompt tokens on T5-Base.}
    \label{fig_t-sne}
\end{figure*}
\ \\ \noindent
\textbf{Effect of Prompt Length.}
We conduct analyses using the RTE and WiC datasets within the SuperGLUE benchmark. To understand the impact of prompt length on the LAMP's performance, we maintained an "intrinsic rank" $r$ of 8 for the soft prompt on the T5-Base model, varying the prompt lengths $\in\{20, 100, 200\}$. Figure \ref{fig_rank_length}(c) and Figure \ref{fig_rank_length}(d) illustrates that LAMP consistently outperforms other baselines at prompt lengths of 20, 100, and 200. LAMP achieves optimal performance when the prompt length is set to 100, consistent with previous findings that 100 is the optimal hyperparameter for prompt length \citep{lester2021power,razdaibiedina-etal-2023-residual}. The experimental details of other datasets in the Appendix \ref{app_Rank+length}.
\ \\ \noindent
\textbf{Impact of Model Scale.} 
Based on quantification, we conducted experiments on T5-11B and Llama2-7B using randomly selected datasets and compared LAMP against baselines that initialize prompts with semantic knowledge from samples. As shown in Figure \ref{Fig_11b_llama}(a) and Figure \ref{Fig_11b_llama}(b), across different model architectures with more than 7B parameters (T5 with an encoder-decoder structure and Llama2 with a decoder-only structure), LAMP consistently helps models adapt to various downstream tasks, achieving superior performance. The experimental details of T5-3B can be found in the Appendix \ref{app_T5-3B}. Meanwhile, Appendix \ref{app_Parameters} details the changes in training parameters and memory usage associated with the model scale.
\ \\ \noindent
\textbf{Blocks of Average Pooling.} 
Figure \ref{Fig_11b_llama}(c) illustrates the changes in model training time and performance on the SuperGLUE benchmark as the number of average pooling blocks increases. The larger the pooling block $p$, the shorter the prompt length input to the model, resulting in a more noticeable reduction in training time. Furthermore, we were pleasantly surprised to find that average pooling had minimal impact on performance, yet it provided significant advantages for PT. This finding offers a promising mentality for extending PT to various domains.

\vspace{-13pt}
\subsection{Interpretability (RQ3)}
\label{interpretability}
Figure \ref{fig_t-sne} compares general PT before (i.e., blue points) and after (i.e., red points) extracting the intrinsic semantic associations between soft prompt tokens on the MultiRC and COPA datasets. To more intuitively reflect the discreteness of PT, we did not standardize the dimensions for comparison. The more extensive the x and y-axis ranges, the higher the degree of discreteness. The features extracted by LAMP exhibit more distinct clustering, indicating LAMP's superior ability to capture and represent the intrinsic structure and patterns of the data. By considering the interactions between prompt tokens, LAMP uncovers the intrinsic semantic associations among tokens to enhance knowledge representation. This enables PLMs to better grasp the semantic content of textual data. The comparative results of discreteness between LAMP and original prompt tuning (PT) across other datasets are detailed in Appendix \ref{app_visualization}.

\section{Related Work}
\textbf{Parameter-efficient Fine-tuning.} 
Parameter-efficient fine-tuning achieves strong results by training a small subset of parameters, thereby reducing computational costs and improving efficiency. AdapterDrop \citep{ruckle2020adapterdrop} improves efficiency by removing unimportant adapters for a given task in each layer of the Transformer. BitFit \citep{zaken2021bitfit} only updates the bias terms while freezing most of the pre-trained model's parameters.
LST \citep{sung2022lst} reduces training memory by running a small ladder network alongside the pre-trained network. 
LoRA \citep{hu2021lora} re-parameterizes incremental matrices through simple low-rank decomposition. KronA \citep{edalati2022krona} replaces the low-rank decomposition in LoRA with Kronecker product decomposition;  PISSA \citep{meng2024pissa} initializes the low-rank matrices with the weights of the pre-trained model, enhancing performance and efficiency. 
However, prompt tuning \citep{lester2021power} stands out from the rest by achieving good results with training very few parameters.
\ \\ \noindent
\textbf{Prompt-based Fine-tuning.} 
Unlike other PEFT methods, prompt-based fine-tuning methods sustain a controlled increase in trainable parameters, even with substantial model scaling.
Prompt tuning \citep{lester2021power} only adds the soft prompt to the input embedding layer of the model.
DPT\citep{xiao2023decomposed} employs a re-parameterization strategy, using two low-rank matrices to replace the original soft prompt. 
DPT relies solely on random number generation for the soft prompt, resulting in weaker generalization and higher sensitivity to initialization.
DePT\citep{shi2024dept} decomposes the soft prompt into shorter prompts and pairs of low-rank matrices, which are then used to update the model's weights. 
EPT \citep{lan2024efficient} leverages multi-space projection and prompt fusion to refine soft prompt knowledge, enhancing flexibility and balancing accuracy with computational efficiency for diverse tasks.
Nevertheless, these PT-based methods need more efficiency and task-specific knowledge richness when dealing with long soft prompt. LAMP provides the ability to tailor prompts more precisely and effectively to the specific requirements of various tasks.
\section{Conclusions}
In this work, we observed that soft prompt tokens initialized randomly from the vocabulary lack intrinsic semantic associations to enhance knowledge representation. Additionally, PT-based methods face challenges balancing knowledge richness and computational cost in different tasks. Based on these issues, we approximate soft prompt by proposing a Low-parameter Prompt Tuning (LAMP) method, which utilizes two singular vectors and singular values. LAMP facilitates semantic knowledge interaction, allowing the soft prompt to incorporate more task-specific knowledge.  It can serve as an efficient plugin for various PT-based tasks. Experimental results across three model scales (T5-Small, T5-Base, T5-Large) demonstrate that LAMP achieves high effectiveness and robustness with fewer trainable parameters.

\section*{Acknowledgements}
This work is partially supported by the National Natural Science Foundation of China under Grant (No. 62032013, 62102074), the Science and Technology projects in Liaoning Province (No. 2023JH3/10200005).

\section*{Limitations}
While our method significantly reduces trainable parameters in NLP, its potential applications extend far beyond this domain. Its evaluation in areas beyond NLP and with other advanced large language models remains a work in the future.  While our approach significantly reduces trainable parameters, further quantification of model parameters will also be explored in future research. The intrinsic semantic interactions between soft prompt tokens can be more effectively mined without increasing the number of trainable parameters. In future work, we will explore methods to enhance knowledge representation for PT.

\bibliography{reference}

\appendix

\begin{table*}
  \centering
  \renewcommand{\arraystretch}{1}
  \setlength{\tabcolsep}{0.1cm}
    \begin{tabular}{lrrlll}
    \toprule
    \multicolumn{6}{c}{\textbf{SuperGLUE Benchmark}} \\
    \midrule
    \textbf{Dataset} & \multicolumn{1}{l}{\textbf{\#Train}} & \multicolumn{1}{l}{\textbf{\#Dev}} & \textbf{Type} & \textbf{Domain} & \textbf{\#Metric} \\
    \midrule
    CB    & 250   & 56    & Natural Language Inference  & various & accuracy  \\
    WSC   & 259   & 104    & Common Sense Reasoning & fiction books & accuracy  \\
    COPA   & 400 & 100   & Question Answering & blogs, etc. & accuracy  \\
    RTE   & 2,490 & 277   & Natural Language Inference   & News, Wikipedia & accuracy  \\
    Wic   & 5,428 & 638   & Word Sense Disambiguation & lexical databases & accuracy  \\
    BoolQ & 9,427 & 3,270 & Question Answering & Wikipedia & accuracy  \\
    MulticRC & 27,243 & 4,848 & Question Answering & various & F1\\
    ReCoRD & 100,730 & 10,000 & Common Sense Reasoning & news (CNN, Daily Mail) & 
    F1\\
    \midrule
    MNLI  & 392,702 & 19,647 & NLI   & various & accuracy  \\
     QNLI  & 103,743 & 6,463 & NLI   & Wikipedia & accuracy  \\
     SST-2 & 66,349 & 1872   & Sentiment & Movie Reviews & accuracy \\
     MRPC  & 3,668 & 408   & Paraphrase & news  & 
    accuracy \\
    \bottomrule
    \end{tabular}%
    \caption{The details of the 8 datasets in SuperGLUE benchmark utilized in our experiment.}
  \label{tab_datasets}%
\end{table*}%

\section{Appendix}
\label{sec:appendix}

\subsection{Self-Attention Pooling}
\label{app_Self-Attention Pooling}
We propose utilizing a self-attention mechanism for the pooling operation to allow the soft prompt to adaptively assign different weights to prompt tokens —emphasizing key tokens while ignoring less important ones. The formula is expressed as follows:
\begin{eqnarray}
\begin{aligned}
    \mathbf{K} & =\mathbf{C} W_{s a} \\
    \mathbf{A}_{weight} & =\operatorname{Softmax}(\mathbf{K}) \\
\mathbf{P'} & =\mathbf{A}_{weight}^{\top} \mathbf{C}
\end{aligned}
\end{eqnarray}
where, $W_{sa} \in \mathbb{R}^{d \times l/p}$ is a learnable initialization weight matrix, and $A_{weight}\in \mathbb{R}^{l \times l/p}$ represents the attention weights. By applying $A_{weight}$ to $\mathbf{C} \in \mathbb{R}^{l \times d}$, we achieve adaptive selection and pooling operation of prompt tokens, resulting in $\mathbf{P'} \in \mathbb{R}^{l/p \times d}$. The self-attention pooling operation introduces additional trainable parameters $W_{sa} \in \mathbb{R}^{d \times l/p}$, and its performance is suboptimal. We hypothesize that this may be due to the compressed outer product effectively capturing the intrinsic semantic associations between prompt tokens. The dynamic weight assignment might disrupt these previously captured associations.

\subsection{Dataset Details}
\label{app_Dataset Details}
Table \ref{tab_datasets} provides detailed information on the 8 datasets we used in SuperGLUE benchmark. The processing of all datasets follows the approach of \citet{xiao2023decomposed}.
%

\begin{table*}[t]
  \renewcommand{\arraystretch}{1}
  \setlength{\tabcolsep}{0.2cm}
  \centering
    \begin{tabular}{cccccccccc}
    \toprule
    \textbf{Model} & \textbf{Params.} & CB    & WSC   & COPA  & RTE   & WiC   & BoolQ & MultiRC & ReCoRD \\
    \midrule
    \multicolumn{10}{c}{\textbf{T5-Small}} \\
    \textbf{LAMP\_S} & 6K    & 1.54  & 2.26  & 0.47  & 0.78  & 0.71  & 0.23  & 1.12  & 0.25 \\
    \midrule
    \multicolumn{10}{c}{\textbf{T5-Base}} \\
    \textbf{LAMP\_B} & 9K    & 0.84  & 2.15  & 1.25  & 0.51  & 0.34  & 0.13  & 0.32  & 0.05 \\
    \midrule
    \multicolumn{10}{c}{\textbf{T5-Large}} \\
    \textbf{LAMP\_L} & 11K   & 1.37  & 2.48  & 0.47  & 0.31  & 0.29  & 0.07  & 0.08  & 0.07 \\
    \bottomrule
    \end{tabular}%
    \caption{We report standard deviation of three runs for our method LAMP, where \text{\_S} is T5-Small, \text{\_B} is T5-Base and \text{\_L} is T5-Large.}
  \label{tab:Standard Deviation}%
\end{table*}%

\begin{table*}[t]
  \centering
  \renewcommand{\arraystretch}{1}
  \setlength{\tabcolsep}{0.45cm}
 
    \begin{tabular}{l|l|cccc|c}
    \toprule
     \multirow{2}[2]{*}{\textbf{Model}} & \multicolumn{1}{l|}{\multirow{2}[2]{*}{\textbf{Param.}}} & \textbf{MNLI} & \textbf{QNLI} & \textbf{SST-2} & \textbf{MRPC} & \textbf{Mean} \\
          &       & 393K  & 105K  & 67K   & 3.7K  & (\%) \\
    \midrule
    Fine-Tuning & 220M  & 86.33  & 93.19  & 94.75  & 84.56  & 89.71  \\
    LoRA  & 3.8M  & 85.30  & 92.96  & 94.04  & 68.38  & 85.17  \\
    PiSSA & 3.8M  & 85.75  & 93.15  & 94.07  & 76.31  & 87.32  \\
    rsLoRA & 3.8M  & 85.73  & 93.12  & 94.19  & 52.86  & 81.48  \\
    LoRA+  & 1.6M  & 85.81  & 93.14  & 93.85  & 74.43  & 86.81  \\
    DoRA  & 3.8M  & 85.67  & 93.04  & 94.04  & 68.08  & 85.21  \\
    LoRA-GA  & 3.8M  & 85.70  & 93.18  & 94.11  & 85.29  & 89.57  \\
    \midrule
    DEPT  & 76.8K  & 85.12  & 93.20  & 94.19  & 88.71   & 90.31  \\
    EPT  & 76.8K  & 85.63  & 93.15  & 94.21  & 89.20  & 90.55  \\
    DPT  & 9K  & 85.34  & 93.15 & 94.50  & 88.47  & 90.37  \\
    \midrule
    \rowcolor{gray!25}
    \textbf{LAMP}  & 7K    & \textbf{85.82 } & \textbf{93.32 } & \textbf{94.50 } & \textbf{90.20 } & \textbf{90.96 } \\
    \bottomrule
    \end{tabular}%
    \caption{The performance comparison between LAMP and the latest LoRA-based PEFT methods on the GLUE benchmark, all experimental results are based on the T5-Base model. The LoRA-based baseline results are derived from LoRA-GA \citep{wang2024loraGA}.  }
  \label{tab:lora}%
  \vspace{-10pt}
\end{table*}%

\begin{table*}[t]
\renewcommand{\arraystretch}{1}
\setlength{\tabcolsep}{0.04cm}
  \centering
    \begin{tabular}{c|c|cccccccc|>{\columncolor{gray!25}}c}
    \toprule
    \multicolumn{1}{c}{\multirow{2}[3]{*}{\textbf{K-Shot}}} & \multirow{2}[3]{*}{\textbf{Method}} & \multicolumn{9}{c}{\textbf{SuperGLUE}} \\
\cmidrule{3-11}    \multicolumn{1}{c}{} &       & \textbf{CB} & \textbf{WSC} & \textbf{COPA} & \textbf{RTE} & \textbf{WiC} & \textbf{BoolQ} & \textbf{MultiRC} & \textbf{ReCoRD} & \textbf{Average} \\
 \midrule
    \multirow{7}[1]{*}{8} & PT    & 58.57  & 32.69  & 40.66  & 49.64  & 53.61  & 53.94  & 50.17  & 46.55  & 48.23  \\
          & Res PT & 60.55    & 28.95    & 55.00    & 47.29    & 44.51    & 61.35    & 57.78    & 68.13    & 52.95 \\
          & DePT  & 61.43  & 36.69  & 56.66  & 48.92  & 51.72  & 47.83  & 47.45  & 47.94  & 49.83  \\
          & EPT   & 57.14  & 42.31  & 45.33  & 51.80  & 50.47  & 53.39  & 56.91  & 46.04  & 50.42  \\
          & DPT   & 62.50  & 39.47  & 55.66  & 56.68  & 50.00  & 62.17  & 60.12  & 63.49  & 56.26  \\
          \cmidrule{2-11}
          & \textbf{LAMP} & 62.86$_{1.53}$  & 44.74$_{1.24}$  & 59.00$_{0.94}$  & 53.43$_{0.2}$  & 50.00$_{0.00}$  & 62.17$_{0.26}$  & 59.98$_{0.00}$  & 62.35$_{1.15}$ & \textbf{57.21}\\
    \midrule
    \multirow{7}[2]{*}{16} & PT    & 57.14  & 32.69  & 44.00  & 51.08  & 53.29  & 58.10  & 56.02  & 46.34  & 49.83  \\
          & Res PT & 68.88    & 50.00    & 55.00    & 47.65    & 53.13    & 62.42    & 55.24    & 68.22    & 57.57 \\
          & DePT  & 50.00  & 37.69  & 48.66  & 55.40  & 49.53  & 57.98  & 52.43  & 50.77  & 50.31  \\
          & EPT   & 42.86  & 42.31  & 48.00  & 51.80  & 56.11  & 57.43  & 57.73  & 49.47  & 50.71  \\
          & DPT   & 44.64  & 55.26  & 55.66  & 53.07  & 51.41  & 62.17  & 59.99  & 62.57  & 55.60  \\
          \cmidrule{2-11}
          & \textbf{LAMP} & 62.21$_{2.66}$  & 47.37$_{2.48}$  & 59.00$_{1.89}$  & 54.51$_{0.85}$  & 54.23$_{0.59}$ & 62.29$_{0.06}$  & 59.88$_{0.01}$  & 70.72$_{0.44}$  & \textbf{58.14 } \\
    \midrule
    \multirow{7}[2]{*}{32} & PT    & 57.14  & 32.69  & 46.33  & 54.68  & 55.49  & 59.94  & 51.39  & 49.15  & 50.85  \\
          & Res PT & 69.21    & 47.37   & 58.66    & 51.99    & 53.34    & 63.09    & 56.15    & 68.15    & 58.50 \\
          & DePT  & 53.57  & 32.69  & 52.33 & 50.36  & 52.98  & 58.72  & 49.89  & 48.91  & 49.93  \\
          & EPT   & 67.85  & 40.38  & 56.33  & 51.80  & 55.80  & 60.12  & 49.74  & 46.05  & 53.51  \\
          & DPT   & 62.50  & 52.63  & 55.00  & 53.79  & 52.35  & 62.17  & 59.95  & 63.34  & 57.72  \\
          \cmidrule{2-11}
          & \textbf{LAMP} & 63.14$_{2.23}$  & 47.37$_{1.54}$  & 60.00$_{1.41}$  & 54.87$_{0.34}$  & 56.74$_{0.44}$  & 62.17$_{0.04}$  & 60.08$_{0.08}$  & 67.80$_{1.71}$  & \textbf{59.25 } \\
    \bottomrule
    \end{tabular}%
  \caption{Few-shot adaptation results (\%) with $k = \{8, 16, 32\}$ on SuperGLUE benchmark. All results are presented as the average of three runs and subscripts indicate standard deviation, each with different random seeds on T5-Base. The
best result is marked in bold.}
  \label{tab:app_few_shot}%
\end{table*}%

\subsection{Baselines Details}
\label{app_Baselines Details}
Apart from full fine-tuning, due to the significant advantages of PT, all other baselines are variants based on PT. Descriptions of all \textbf{PT-based} baselines are as follows:
\begin{itemize}
\item \textbf{Full Fine-tuning:} Updating all model parameters in the T5-models \citep{raffel2020exploring} on each downstream task. It is the most fundamental method for comparing PEFT methods' performance and trainable parameters.
\item\textbf{Prompt tuning} \citep{lester2021power}: PT stands out in the PEFT approaches because it freezes the parameters of PLMs and only trains the attached soft (continuous) prompt vectors to the input text.
\item\textbf{Residual Prompt tuning} \citep{razdaibiedina-etal-2023-residual}: A PT-based variant (named Res PT) that utilizes a residual network to increase the flexibility of model selection for soft prompt token representations and improve the convergence rate.
\item\textbf{DePT} \citep{shi2024dept}: Decomposing the prompt into a shorter prompt and low-rank matrix pairs to reduce training time. It utilizes low-rank matrix pairs to update the input embedding.
\item \textbf{EPT} \citep{lan2024efficient}: This method utilizes multi-space projection and prompt fusion modules to enhance the soft prompt knowledge, making it more adaptable to various downstream tasks while balancing accuracy and efficiency.
\item \textbf{DPT} \citep{xiao2023decomposed}: A novel prompt initialization method involves replacing the prompt with two randomly initialized low-dimensional matrices.
 \end{itemize}
Descriptions of all \textbf{LoRA-based} baselines are as follows:
\begin{itemize}
\item\textbf{LoRA} \citep{hu2021lora}: A parameter-efficient approach focuses on updating only the low-rank matrices within the model.
\item\textbf{PiSSA} \citep{meng2024pissa}: It performs Truncated Singular Value Decomposition (SVD) on the model's weight matrix and uses the resulting low-rank matrices as the initialization for the low-rank matrices $A$ and $B$.
\item\textbf{rsLoRA} \citep{kalajdzievski2023rsLoRA}:  introduces a new scaling factor to stabilize LoRA's parameter scaling.
\item\textbf{LoRA+} \citep{hayoulora+}: It utilizes two different learning rates to control the updates of the low-rank matrices $A$ and $B$.
\item\textbf{DoRA} \citep{liudora}: It decomposes the pre-trained weight into two components, $magnitude$ and $direction$, for fine-tuning.
\item\textbf{LoRA-GA} \citep{wang2024loraGA}: It aligns low-rank gradient products with full fine-tuning gradients at the first step.
 \end{itemize}
\subsection{Implementation Details}
\label{app_Implementation_Details}
Weight decay of $1e-5$, and the maximum sequence length for the model is typically configured at 256. LAMP is implemented by the Python library of PyTorch 2.0.0 \footnote{\footnotesize \url{https://pytorch.org/}}, Huggingface Transformers 4.30.0 \footnote{\footnotesize \url{https://github.com/huggingface/transformers}}. All of our experiments were conducted with 8 GPUs, with 48 GB memory each. 

\subsection{Standard Deviation}
\label{app_Standard Deviation}
We present the standard deviation across three runs for our method on T5-Small, T5-Base and T5-Large. The outcomes are provided in Table \ref{tab:Standard Deviation}.

\subsection{Performance Comparison with LoRA Variants}
\label{app_lora}
Table \ref{tab:lora} presents a performance comparison between LAMP and LoRA, along with its variants on the GLUE benchmark. LAMP outperforms all other novel LoRA-based baselines. Notably, LAMP requires significantly fewer trainable parameters than all baselines, representing a unique advantage among PEFT methods.

\subsection{Few-shot Adaptions Details}
\label{app_Few-shot Details}
In Table  \ref{tab:app_few_shot}, we present the results of all baseline models across all SuperGLUE datasets, using T5-Base as the benchmark. The subscripts represent the standard deviation of our method LAMP across different K-shot settings.

\begin{figure}[htbp]
\centering
  \includegraphics[scale=0.26]{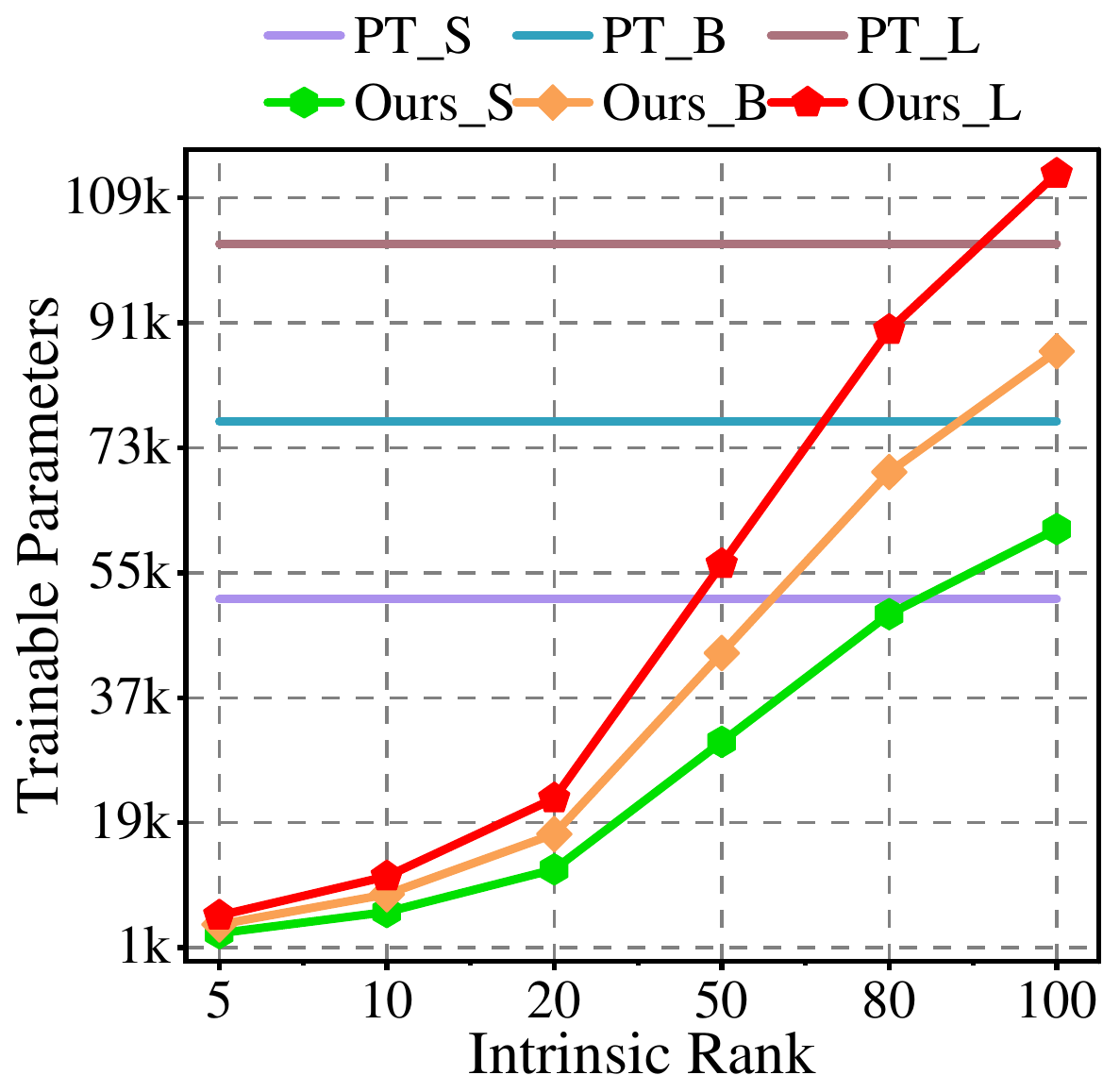}
    \caption{Trainable parameters with different intrinsic ranks on the T5 models, where \text{\_S} is T5-Small, \text{\_B} is T5-Base and \text{\_L} is T5-Large. }
    \label{fig_rank_size}
    \vspace{-10pt}
\end{figure}

\subsection{Change in LAMP Parameters from Intrinsic Rank}
\label{app_Intrinsic Rank}
As shown in Figure \ref{fig_rank_size}, the impact of intrinsic rank $r \in \{5, 10, 20, 50, 80, 100\}$ on LAMP trainabel parameters is illustrated across T5 models (T5-Small, T5-Base, and T5-Large). As $r$ increases, LAMP's trainable parameters gradually approach that of the vanilla PT. When all ranks in the Truncated SVD are retained, the trainable parameters of PT are fewer than those of LAMP. 
We downplay the impact of prompt length on the number of trainable parameters. The number of trainable parameters in LAMP can be further reduced by adjusting $r$.


\begin{figure*}[t]
\centering
  \includegraphics[scale=0.21]{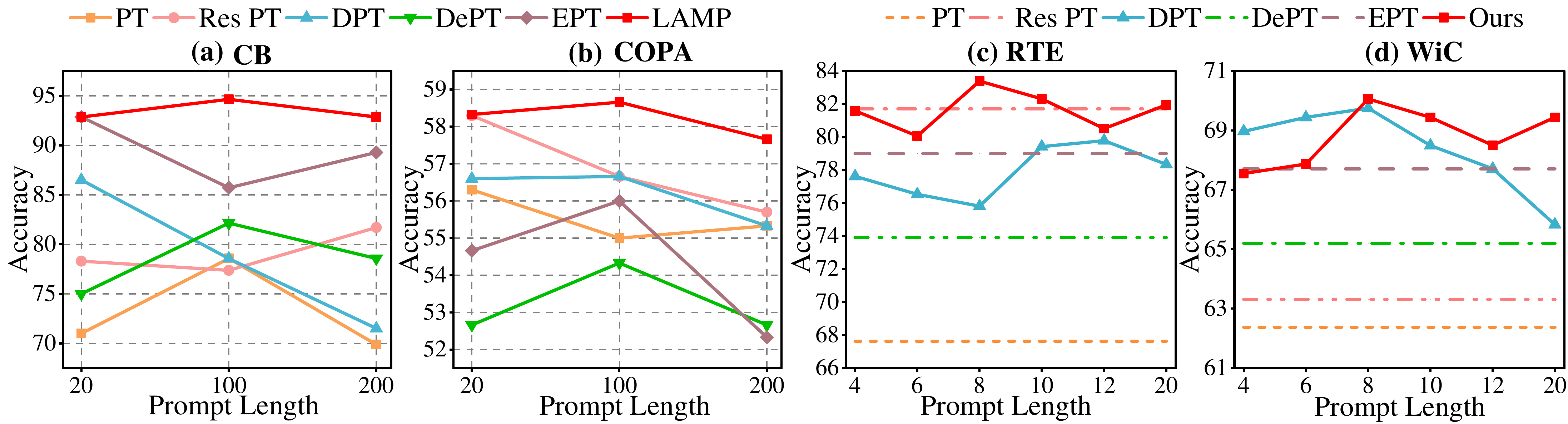}
    \caption{(a) and (b), performance of different baselines varies with the prompt length $l \in \{20, 100, 200\}$ on the CB and COPA datasets. (c) and (d), compare the performance of all baselines with the number of inherent ranks $r \in\{4, 6, 8, 10, 12, 20\}$ on the RTE and WiC datasets.  All results represent the average of three runs conducted with a different random seed.}
    \label{fig_length+rank}
\end{figure*}

\subsection{Impact of Rank Size and Prompt Length}
\label{app_Rank+length}
As shown in Figure \ref{fig_length+rank}, we present the effects of intrinsic rank and prompt length across various datasets. The experimental results indicate that LAMP consistently performs strongly across different intrinsic ranks $r$ and prompt lengths $l$.

\subsection{Performance in T5-3B}
\label{app_T5-3B}
For baseline selection, we similarly chose baselines initialized from sampled vocabulary. Table \ref{tab_T5-3B} shows that LAMP performs best on the T5-3B model, further validating the importance of incorporating intrinsic semantic associations between prompt tokens to enhance knowledge representation capacity.
\begin{table}[t]
  \centering
\renewcommand{\arraystretch}{1}
\setlength{\tabcolsep}{0.25cm}
\begin{tabular}{l|cccc}
    \toprule
    \multirow{2}[2]{*}{\textbf{Method}} & \multicolumn{4}{c}{\textbf{T5-3B}} \\
    \cline{2-5}
          & PT   & DEPT   & EPT   & LAMP  \\
    \midrule
    MultiRC & 78.00& 77.95 & 78.51 & 78.83 \\
    WiC & 71.16 & 71.16 & 73.67 & 74.92 \\
    RTE & 91.31 & 92.75 & 92.75 & 93.48 \\
    BoolQ & 87.22 & 87.65 & 87.89 & 87.83 \\
    \midrule
    \rowcolor{gray!25}
    Mean & 81.93 & 82.38 & 83.21 & 83.77 \\
    \bottomrule
    \end{tabular}%
\caption{The performance changes of different
methods at various datasets on the T5-3B. }
  \label{tab_T5-3B}%
\end{table}%

\subsection{Change in LAMP Parameters}
\label{app_Parameters}

\begin{table*}[t]
  \renewcommand{\arraystretch}{1}
  \setlength{\tabcolsep}{0.1cm}
  \centering
    \begin{tabular}{c|ccccc|ccccc}
    \toprule
    \multirow{2}[2]{*}{\textbf{Method}} & \multicolumn{5}{c|}{\textbf{T5-Large}   Prompt Length} & \multicolumn{5}{c}{\textbf{Llama2-7B}  Prompt Length} \\
          & 20    & 100   & 1000  & 5000  & 10000 & 20    & 100   & 1000  & 5000  & 10000 \\
    \midrule
          & \multicolumn{5}{c|}{ \#Trainable Params  } & \multicolumn{5}{c}{ \#Trainable Params} \\
    \midrule
    Full FT & 0.7B   & 0.7B  & 0.7B   & 0.7B   & 0.7B   & 7B   & 7B    & 7B    & 7B    & 7B  \\
    PT & 0.02M & 0.1M & 1.02M & 5.12M & 10.24M & 0.08M & 0.41M & 4.10M & 20.48M  & 40.96M \\
    \textbf{LAMP} & \textbf{0.008M} & \textbf{0.009M} & \textbf{0.016M} & \textbf{0.048M} & \textbf{0.088M} & \textbf{0.033M} & \textbf{0.034M} & \textbf{0.041M} & \textbf{0.073M} & \textbf{0.113M} \\
    \midrule
    Ratio & \blue{2.45}  & \blue{11.38}  & \blue{63.21}  & \blue{106.22}  & \blue{116.10}  & \blue{2.49}  & \blue{12.20}  & \blue{100.45}  & \blue{281.41}  & \blue{363.20}  \\
    \bottomrule
    \end{tabular}%
    \caption{Trainable parameters of different baselines varies with the prompt length. ``Ratio" denotes the multiple of trainable parameters in vanilla prompt tuning (PT) relative to LAMP.}
  \label{tab3}%
\end{table*}%

\subsubsection{Change in LAMP Parameters from Model Size}
\begin{figure}[t]
\centering
  \includegraphics[scale=0.26]{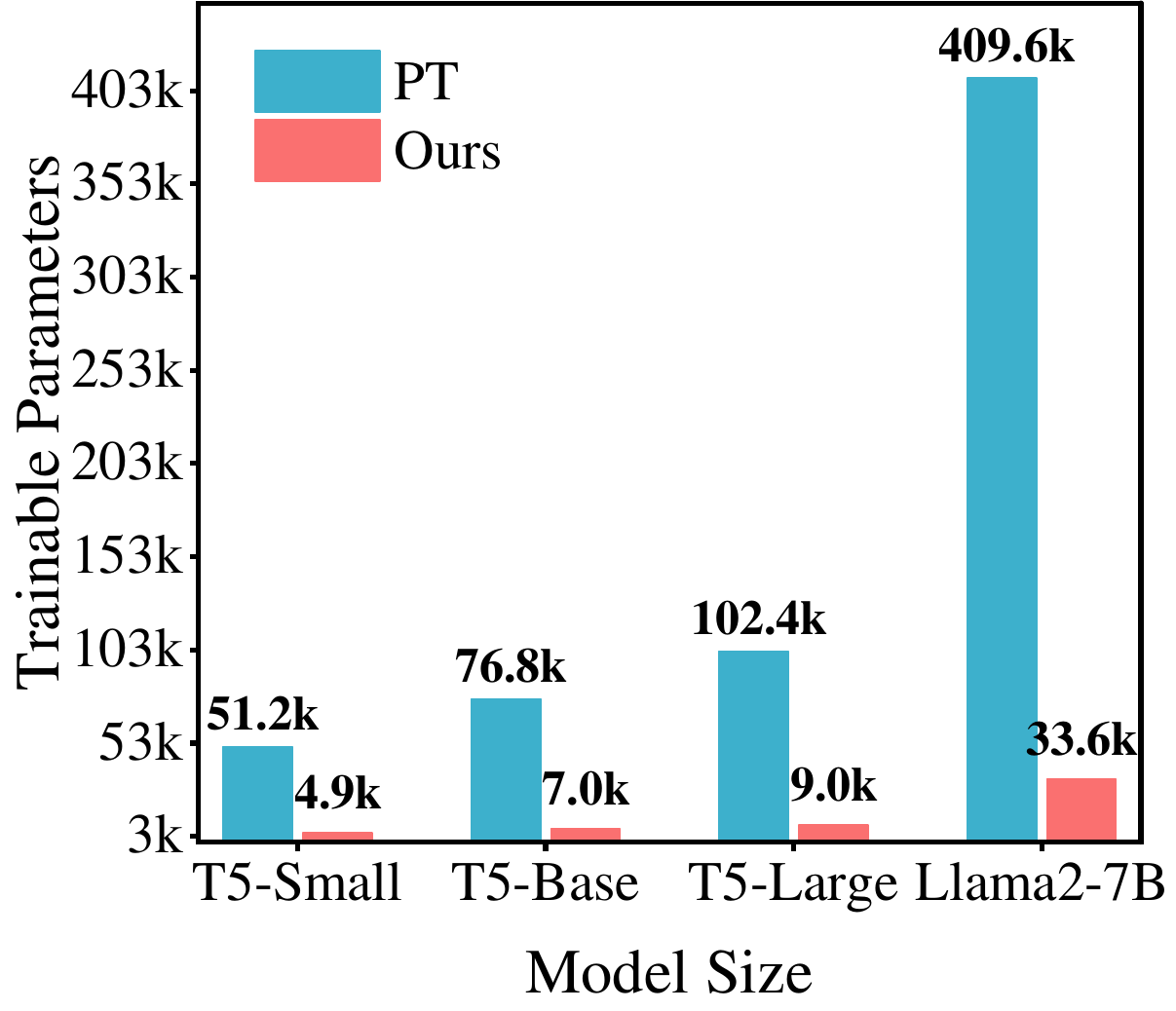}
    \caption{Trainable parameters with different model size. }
    \label{fig_model_size}
    \vspace{-10pt}
\end{figure}

With intrinsic rank $r=8$, Figure \ref{fig_model_size} illustrates the effect of model size on LAMP trainable parameters. As the model size increases, trainable parameters in original PT increases significantly, whereas LAMP mitigates this issue. Table \ref{tab3} shows that when the prompt length $l$ and  intrinsic rank $r$ remain constant, the higher the model size, the more significant the reduction in trainable parameters achieved by LAMP. LAMP also mitigates the impact of the hidden dimension on trainable parameters.

\begin{figure*}[t]
\centering
  \includegraphics[scale=0.21]{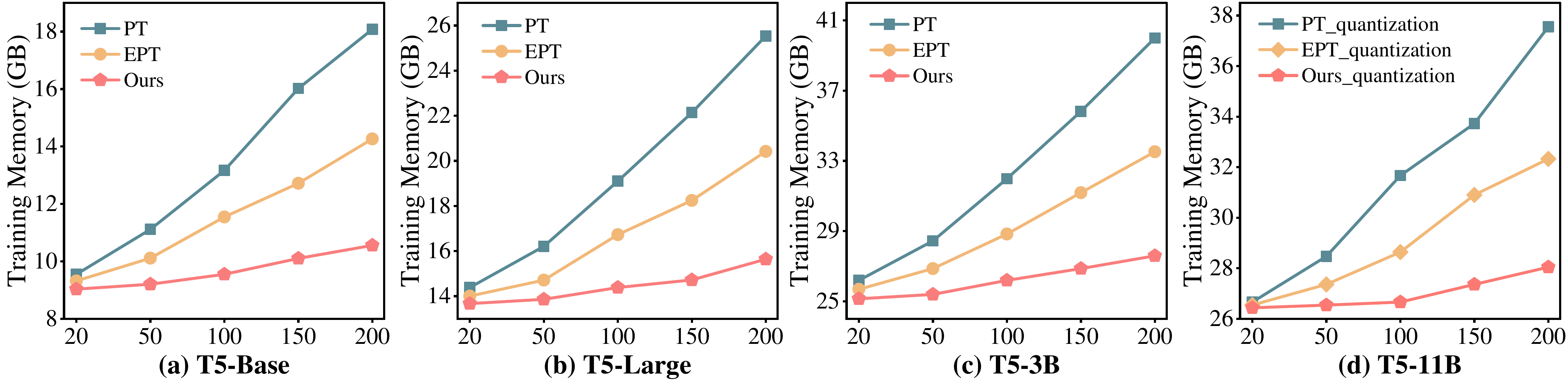}
    \caption{Comparison of memory usage using different methods on various model scales. We leverage quantization operation on T5-11B.}
    \label{fig_memory}
\end{figure*}

\begin{figure*}[t]
\centering
  \includegraphics[scale=0.65]{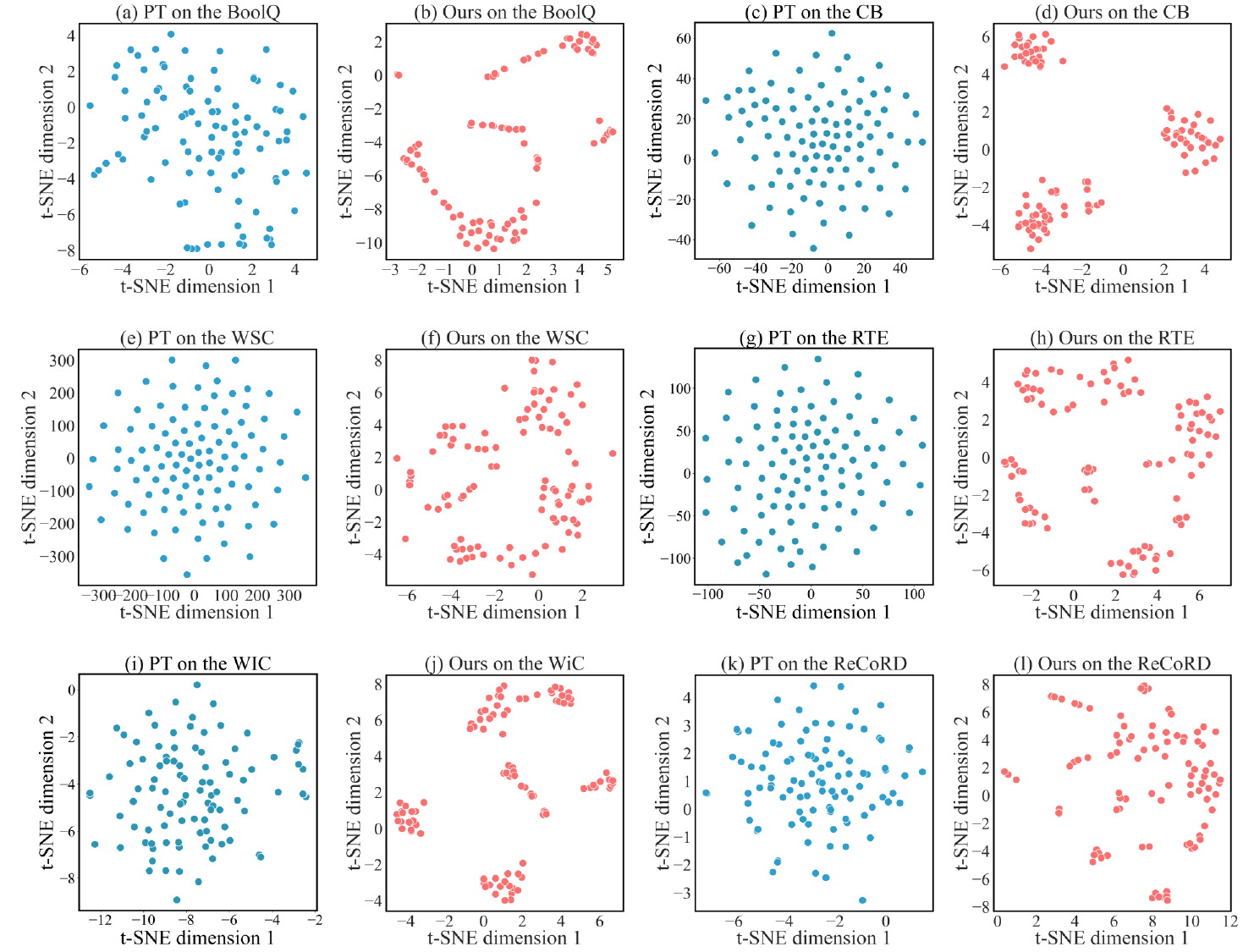}
    \caption{The comparison of the dispersion of PT tokens on the other datasets in SuperGLUE benchmark before and after considering the intrinsic semantic associations between soft prompt tokens on T5-Base.}
    \label{fig_visual}
\end{figure*}

\subsubsection{Variation of LAMP Parameters}
\label{subsec:r3}
Table \ref{tab3} presents the variation of LAMP's trainable parameters with prompt length $l \in \{20, 100, 1000, 5000, 10000\}$ with $r=8$. 
The longer the prompt length, the more LAMP downplays the impact of prompt length on trainable parameters, making LAMP's advantages more evident. For instance, in the Llama2-7B model, when the prompt length is set to 10,000, the number of trainable parameters is reduced to just one-three-hundred-sixtieth of the original PT, significantly boosting computational efficiency. 
In summary, influenced by intrinsic rank $r$, prompt length $l$ , and model size, the LAMP's efficiency compared to PT and other PEFT methods can be further expanded. 

\subsubsection{Change in LAMP memory usage from Model Size}
We visualized the memory usage of LAMP across different model scales, with the T5-11B results reflecting quantization. As shown in Figure \ref{fig_memory}, LAMP consistently has the lowest memory usage and highest computational efficiency across various model sizes.

\subsection{Visualization of Intrinsic Semantic Associations}
\label{app_visualization}
We visualized the comparison of general PT before and after extracting the intrinsic semantic associations between soft prompt tokens on other datasets within the SuperGLUE benchmark. Figure \ref{fig_visual} shows that LAMP also exhibits apparent clustering on these datasets, enhancing semantic knowledge representation. 

\end{document}